\documentclass[10pt,journal,cspaper,compsoc]{IEEEtran}
\IEEEoverridecommandlockouts
\usepackage{amsmath,amssymb}
\usepackage{bbm}
\usepackage{mathtools}
\DeclarePairedDelimiter\ceil{\lceil}{\rceil}

\DeclareMathOperator*{\argmax}{arg\,max} 

\usepackage[margin=0.6in]{geometry}

\usepackage{colortbl}
\usepackage{makecell}

\usepackage{graphicx, overpic, wrapfig, subfigure}

\usepackage[dvipsnames]{xcolor}
\definecolor{americanrose}{rgb}{1.0, 0.01, 0.24}
\definecolor{myred}{rgb}{0.753, 0.314, 0.275}
\definecolor{myblue}{rgb}{0.0, 0.24, 0.95}
\definecolor{tbl_gray}{gray}{0.85}

\newcommand\MYhyperrefoptions{bookmarks=true,bookmarksnumbered=true,
pdfpagemode={UseOutlines},plainpages=false,pdfpagelabels=true,
colorlinks=true,linkcolor={americanrose},citecolor={myblue},urlcolor={myblue},
pdftitle={Deep Hough Transform for Semantic Line Detection},
pdfsubject={Typesetting},
pdfauthor={Kai Zhao et al.},
pdfkeywords={Straight line detection, Hough transform, CNN}}
\usepackage[\MYhyperrefoptions,pdftex]{hyperref}

\usepackage{cleveref}
\crefname{equation}{Eq.}{Eq.}
\crefname{figure}{Fig.}{Fig.}
\crefname{table}{Tab.}{Tab.}
\crefname{section}{Sec.}{Sec.}

\newcommand{\revise}[1]{{\textcolor{black}{#1}}}
\newcommand{\rerevise}[1]{{\textcolor{black}{#1}}}

\ifdefined \GramaCheck
  \newcommand{\CheckRmv}[1]{}
  \renewcommand{\eqref}[1]{Equation 1}
\else
  \newcommand{\CheckRmv}[1]{#1}
  \renewcommand{\eqref}[1]{Equation~(\ref{#1})}
\fi

\newcommand{\Arixv}{}
\ifdefined \Arixv
    \newcommand{\ArxivRmv}[1]{}
\else
    \newcommand{\ArxivRmv}[1]{#1}
\fi

\def\etal{\emph{et al.~}}
\def\ie{\emph{i.e.,~}}
\def\eg{\emph{e.g.,~}}

\usepackage{booktabs} 
\usepackage{array}
\usepackage{diagbox}
\usepackage{multirow}
\usepackage{colortbl}

\usepackage{silence}
\graphicspath{{./figures/}{./figures/photos/}}

\usepackage[nocompress]{cite}

\begin{document}

\title{Deep Hough Transform for Semantic Line Detection}

\author{\IEEEauthorblockN{Kai Zhao\thanks{\IEEEauthorrefmark{1} The first two students contribute equally to this paper.}\IEEEauthorrefmark{1}},
        Qi Han\IEEEauthorrefmark{1},
        Chang-Bin Zhang,
        Jun Xu,
        \IEEEauthorblockN{Ming-Ming Cheng\thanks{\IEEEauthorrefmark{2} M.M. Cheng is the corresponding author (cmm@nankai.edu.cn).}\IEEEauthorrefmark{2}},~\IEEEmembership{Senior Member,~IEEE}

\IEEEcompsocitemizethanks{
  \IEEEcompsocthanksitem Kai Zhao, Qi Han, Chang-Bin Zhang, and Ming-Ming Cheng are with the TKLNDST, 
    College of Computer Science, Nankai University, Tianjin, China, 300350.
\IEEEcompsocthanksitem  Jun Xu is with the School of Statistics and Data Science, Nankai University, Tianjin, China, 300071.
  \IEEEcompsocthanksitem A preliminary version of this work has been presented in \cite{eccv2020line}.
}
}
\markboth{IEEE Transactions on Pattern Analysis and Machine Intelligence  \ \ \ \ \ \ \ \ \ \ \ \ \ \ \ \ \ \ \ \ \ \ \ \ \ \ \ \ \ \ \ \ \ \ \ 
\ \ \ \ \ \ \ \ \ \ \ \ \ \ \ \ \ \ \ \ \ \ \ \ \ \ \ \ \ \ \ \ \ \ \  \ \ \ \ \ 
DOI:~\href{https://doi.org/10.1109/TPAMI.2021.3077129}{TPAMI.2021.3077129}}%
{Zhao \MakeLowercase{\textit{et al.}}: 
Deep Hough Transform for Semantic Line Detection}

\IEEEtitleabstractindextext{%
\begin{abstract}
We focus on a fundamental task of detecting meaningful line structures, 
\textsl{a.k.a.}, semantic line, in natural scenes.
Many previous methods regard this problem as a special 
case of object detection
and adjust existing object detectors for semantic line detection.
However, these methods neglect the inherent characteristics of lines, 
leading to sub-optimal performance.
Lines enjoy much simpler geometric property than complex objects
and thus can be compactly parameterized by a few arguments.
To better exploit the property of lines, in this paper, 
we incorporate the classical Hough transform technique into 
deeply learned representations and 
propose a one-shot end-to-end learning framework for line detection.
%
By parameterizing lines with slopes and biases, 
we perform Hough transform to translate 
deep representations into the parametric domain,
in which we perform line detection.
Specifically, we aggregate features along candidate lines on the 
feature map plane 
and then assign the aggregated features to corresponding locations 
in the parametric domain.
Consequently, the problem of detecting semantic lines in the spatial domain 
is transformed into spotting individual points in the parametric domain, 
making the post-processing steps,
\ie non-maximal suppression, more efficient.
Furthermore, our method makes it easy to extract contextual line features
that are critical for accurate line detection.
In addition to the proposed method, we design an evaluation metric to assess the
quality of line detection and construct a large scale dataset for the line detection task.
Experimental results on our proposed dataset and another public dataset 
demonstrate the advantages of our method over previous state-of-the-art 
alternatives.
The \revise{dataset and} source code is available at~\url{https://mmcheng.net/dhtline/}.
\end{abstract}

}

\maketitle

\IEEEdisplaynontitleabstractindextext

\IEEEpeerreviewmaketitle

\section{Introduction}\label{sec:introduction}

\IEEEPARstart{D}etecting line structures from digital images has 
a long history in computer vision.
The organization of line structures is an early yet essential step to transform the visual signal into useful intermediate concepts 
for visual interpretation~\cite{burns1986extracting}.
%
Though many techniques have been proposed to detect salient objects
\cite{zhao2019optimizing,HouPami19Dss,gao2020sod100k,BorjiCVM2019,wang2021revisiting}
and areas~\cite{cheng2015global,zhu2014saliency,Fan2020S4Net},
little work has been made for detecting outstanding/structure-revealing line structures.
A recent study~\cite{lee2017semantic} was proposed to detect outstanding straight line(s),
referred to as ``semantic line'', that outlines the conceptual structure of natural images.
%
Identifying these semantic lines is of crucial importance for 
computer graphics and vision applications, such as 
photographic composition~\cite{liu2010optimizing,freeman2007photographer},
structure-preserving image processing~\cite{TIP20_SP_NPR,hu2013patchnet},
image aesthetic~\cite{ko2018pac,lee2019property,kong2016photo,mai2016composition},
lane detection~\cite{fan2019spinnet},
and artistic creation~\cite{krages2012photography,hu2013inverse,chen2009sketch2photo,zhang2020and}.
\revise{
  As demonstrated in~\cref{fig:photographic-composition}, 
Liu \etal~\cite{liu2010optimizing} proposed to crop images according 
to the golden ratio by using `prominent line'.
Detecting these `semantic lines' can help to produce images that
are visually pleasing in the photographic composition.
}

The Hough transform~\cite{duda1971use,ballard1981generating} is one 
representative method for line detection,
which was first proposed to detect straight lines in bubble chamber 
photographs~\cite{hough1962method}.
Since its simplicity and efficiency, 
HT is employed to detect lines in digital images~\cite{duda1971use},
and further extended by~\cite{ballard1981generating} to detect other regular shapes like circles and rectangles.
The key idea of the Hough transform is to vote evidence from the image domain to the parametric domain, and
then detect shapes in the parametric domain by identifying local-maximal responses.
In the case of line detection, a line in the image domain can be represented 
by its parameters, \eg slope, and \revise{offset} in the parametric space.
Hough transform collects evidence along with a line in an image
and accumulates evidence to a single point in the parameter space.
Consequently, line detection in the image domain is converted to the problem of detecting peak responses in the
parametric domain.
Classical Hough transform based line detectors
~\cite{fernandes2008real,yacoub1995hierarchical,princen1990hierarchical,kiryati1991probabilistic}
usually detect continuous straight edges while neglecting the semantics in line structures.
Moreover, these methods are sensitive to light changes and occlusion.
Therefore, the results are often noisy and contain irrelevant lines~\cite{akinlar2011edlines},
as shown in~\cref{fig:photographic-composition}(d).

\newcommand{\addImg}[1]{\subfigure[]{\includegraphics[width=.245\linewidth]{figures/#1}}}
\CheckRmv{
\begin{figure*}[t]
  \centering
  \hfill
  \addImg{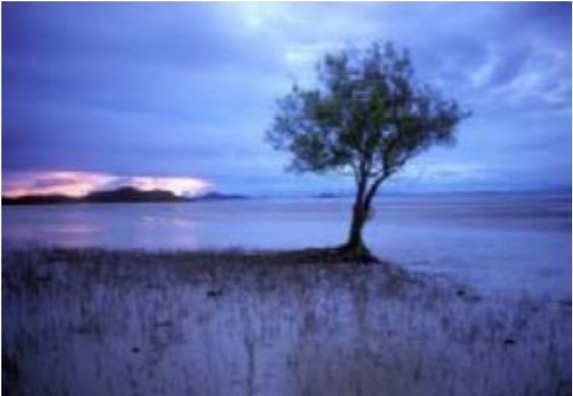}
  \addImg{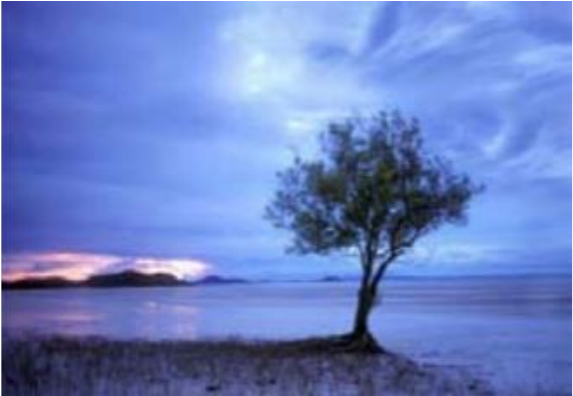}
  \addImg{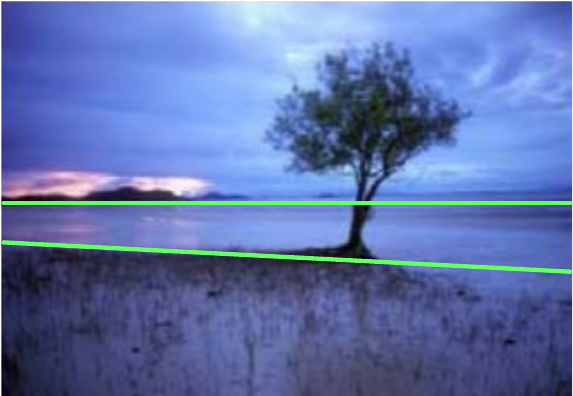}
  \addImg{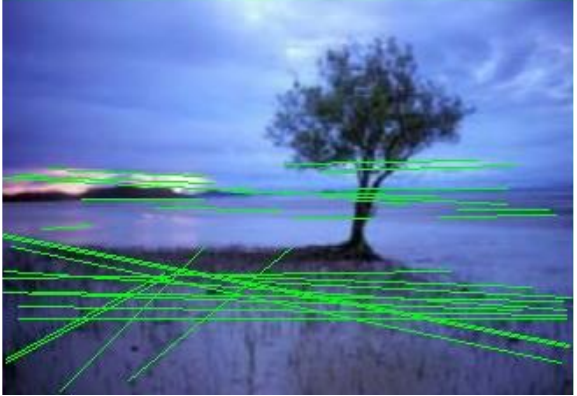}
  \hfill
  \caption{
    Example pictures from~\cite{liu2010optimizing} reveal that semantic lines
    may help in the photographic composition.
    (a): a photo was taken with an arbitrary pose.
    (b): a photo fits the golden ratio principle~\cite{caplin2008art,krages2012photography} which \revise{is} obtained by the method described
    in\cite{liu2010optimizing} using so-called `prominent lines' 
    in the image.
    (c): Our detection results \revise{are} clean and comprise only a few meaningful
        lines that are potentially helpful in the photographic composition.
    (d): Line detection results by the classical line detection algorithms 
    often focus on fine detailed straight edges.
  }
  \label{fig:photographic-composition}
\end{figure*}
}

Convolutional Neural Networks (CNNs) have achieved remarkable success in a wide range of computer vision tasks.
Several recent studies~\cite{lee2017semantic,zhang2019ppgnet} have proposed CNN-based methods for line detection.
Concretely, they regard line detection as a special case of object detection and employ existing object detectors
\eg faster R-CNN~\cite{ren2015faster} or CornerNet~\cite{law2018cornernet},
for line detection.
Limited by the ROI pooling and non-maximal suppression of lines,
both~\cite{lee2017semantic} and~\cite{zhang2019ppgnet} are less efficient in terms of
running time.
%
%
Moreover, ROI pooling~\cite{girshick2015fast} aggregates features along with a
single line, while many recent studies reveal that richer context information is critical to many
tasks, \eg video classification~\cite{wang2018non} and semantic segmentation~\cite{huang2019ccnet}.
This point will be validated in~\cref{sec:ablation}, in which we experimentally verify that only aggregating features along a single line will produces sub-optimal results.

Incorporate powerful CNNs to Hough transform is a promising direction for
semantic line detection.
A simple way of combining CNN with Hough transform is performing edge detection 
with a CNN-based edge detector~\cite{RcfEdgePami2019,xie2015holistically}
and then apply standard Hough transform to the edge maps.
However, the two components have diverse optimization targets, leading to sub-optimal results, as
evidenced by our experiments.
In this paper, we propose to incorporate CNN with Hough transform into an end-to-end manner
so that each component in our proposed method shares the same optimization target.
Our method first extracts pixel-wise representations with a CNN-based encoder
and then performs Hough transform on the deep representations to convert representations
from feature space into parametric space.
Then the global line detection problem is converted to simply detecting peak response in
the transformed features, making the problem much simpler.
For example, the time-consuming non-maximal suppression (NMS) can be simply
replaced by calculating the centroids of connected areas in the parametric space, 
making our method very efficient that can detect lines in real-time.
Moreover, in the detection stage, we use several convolutional layers on
top of the transformed features to aggregate context-aware features of nearby lines.
Consequently, the final decision is made upon not only features of a single line,
but also information about lines nearby.
As shown in \cref{fig:photographic-composition}(c), 
our method detects clean, meaningful and outstanding lines,
that are helpful to photographic composition.

To better evaluate line detection methods, we introduce a principled metric to assess
the agreement of a detected line \textsl{w.r.t.} its corresponding ground-truth line.
Although~\cite{lee2017semantic} has proposed an evaluation metric that uses intersection areas to measure the similarity between a pair of lines,
this measurement may lead to ambiguous and misleading results.
And at last, we collect a large scale dataset with \revise{6,500} carefully annotated images for semantic
line detection.
The new dataset, namely \revise{NKL (short for \textbf{N}an\textbf{K}ai \textbf{L}ines),}
contains images of diverse scenes,
and the scale is much larger than the existing SEL~\cite{lee2017semantic} dataset
in both terms of images and annotated lines.

The contributions of this paper are summarized below:
\begin{itemize}\setlength\itemsep{0.3em}
  \item We proposed an end-to-end framework for incorporating the feature learning capacity of CNN with Hough transform, resulting in an efficient real-time solution for semantic line detection.
  
  \item To facilitate the research of semantic line detection, we construct
    a new dataset with \revise{6,500} images, which is larger and more diverse than a previous SEL dataset~\cite{lee2017semantic}.
  
  \item We introduce a principled metric that measures the similarity between two lines. 
    Compared with the previous IOU based metric~\cite{lee2017semantic},
    our metric has straightforward interpretation and \revise{simplicity in implementation},
    as detailed in~\cref{sec:metric}.
  
    \item Evaluation results on an open benchmark demonstrate that our method 
    outperforms prior arts with a significant margin.
\end{itemize}

A preliminary version of this work was presented in~\cite{eccv2020line}.
In this extended work, we introduce three major improvements:
\begin{itemize}\setlength\itemsep{0.3em}
  \item We propose a novel ``edge-guided refinement'' module to adjust line positions and obtain better detection performance with the help of accurate edge information.
  This part is detailed in~\cref{sec:refine}.
  \item We introduce a new large-scale dataset for semantic line detection, as presented in~\cref{sec:nkl-dataset}.
  The new dataset, \revise{namely NKL
  (short for \textbf{N}an\textbf{K}ai \textbf{L}ines),
  contains 6,500 images in total}, and each image is annotated by multiple skilled annotators.
  \item We employ the maximal bipartite graph matching~\cite{kuhn1955hungarian} to match ground-truth and detected lines during evaluation (\cref{sec:protocol}).
  The matching procedure removes redundant true positives so that each ground-truth line is associated with
  at most one detected line and vice versa.
\end{itemize}

The rest of this paper is organized as follows:
\cref{sec:related-work} summarizes the related works.
\cref{sec:dht} elaborates the proposed Deep Hough transform method.
\cref{sec:metric} describes the proposed evaluating metric, which is used to assess the similarity between a pair of lines.
\cref{sec:nkl-dataset} introduces our newly constructed dataset.
\cref{sec:experiments} presents experimental details and report comparison results.
\cref{sec:conclusion} makes a conclusion remark.

\section{Related Work}\label{sec:related-work}
The research of line detection in digital images dates back to the very early stage of
computer vision research.
Here, we first brief the evolution of Hough transform~\cite{duda1971use} (HT),
one of the most fundamental tools, for line detection.
Then we introduce several recent CNN based methods for line detection.
At last, we summarize the methods and datasets for semantic line detection.

\subsection{Hough transform}
The Hough transform (HT) was firstly proposed in~\cite{hough1962method} for machine analysis of bubble chamber photographs.
It parametrizes straight lines with \revise{slope-offset}, leading to an unbounded transform space (since the slope can be infinity).
\cite{duda1971use} extended HT by using angle-radius rather than \revise{slope-offset}
parameters, and is conceptually similar to two-dimensional Radom transform~\cite{radon20051}.
Then Ballard \etal \cite{ballard1981generating} generalized the idea of HT to localize arbitrary shapes, \eg ellipses and circles, from digital images.
For example, by parameterizing with angle and radius, line detection can be performed by voting edge evidence and finding peak response in the finite parametric space.
Typically, with the edge detectors such as Canny~\cite{canny1986computational} and Sobel~\cite{sobel},
the detected lines are the maximal local response points in the transformed parametric space.
\revise{The core idea of HT is used in two recent
works which
parameterize the outputs of CNNs with offsets and orientations
to predict surface meshes~\cite{chen2020bsp} or
convex decomposition~\cite{deng2020cvxnet} of 3D shapes.}

Despite the success of HT on line detection, it suffers from high computational costs and unstable performance.
%
To accelerate the voting of HT, Nahum \etal \cite{kiryati1991probabilistic} 
proposed the ``probabilistic Hough transform'' to randomly pick 
sample points from a line, 
while~\cite{finding1976} using the gradient direction of images to decide 
the voting points.
Meanwhile, the work of \cite{fernandes2008real,limberger2015real} 
employed kernel-based Hough transform to perform hough voting by using 
the elliptical-Gaussian kernel on collinear pixels to boost the original HT.
Besides, John \etal \cite{princen1990hierarchical,yacoub1995hierarchical} partitioned the input image into
hierarchical image patches, and then applied HT independently to these patches.
Illingworth \etal \cite{illingworth1987adaptive} use a coarse-to-fine accumulation and search
strategy to identify significant peaks in the Hough parametric spaces.
\cite{aggarwal2006line} tackled line detection within a
regularized framework, to suppress the effect of noise and clutter corresponding to nonlinear image features.
The Hough voting scheme is also used in many other tasks such as
detecting centroid of 3D shapes in point cloud~\cite{qi2019deep} and finding image correspondence~\cite{min2019hyperpixel}.

\subsection{Line Segments Detection}
Though its robustness and parallelism, Hough transform cannot be directly used for line segments detection,
since it cannot determine the endpoints of line segments.
Probabilistic Hough transform~\cite{kiryati1991probabilistic} uses random sampling in the voting scheme,
and reconstructs line segments by localizing the sample locations.
But this method still prefers long straight lines.
In addition to Hough transform, many other studies have been developed to detect line segments.
Burns \etal \cite{burns1986extracting} used the edge orientation as the guide for line segments extraction.
The main advantage is that the orientation of the gradients can help to discover low-contrast lines
and endpoints.
Etemadi \etal \cite{etemadi1992robust} established a chain from the given edge map and
extracted line segments and orientations by walking over these chains.
Chan \etal \cite{chan1996line} used a quantified edge orientation to search and merge short line segments.
Gioi \etal \cite{von2008lsd} proposed a linear-time line segment detector (LSD) without tuning parameters,
and is used by many subsequent studies~\cite{akinlar2011edlines,akinlar2013edcircles,feng2013automatic}

\subsection{CNN based Line Detection.}
%
Recently, CNNs have brought remarkable improvements in computer vision tasks, and also be applied to line detection.
These methods either focus on straight line detection, \eg semantic line detection~\cite{lee2017semantic,ahmad2017comparison}, or line segments detection, 
\eg wireframe parsing~\cite{zhang2019ppgnet,huang2018learning,zhou2019end,xue2020holistically}.
Lee \etal \cite{lee2017semantic} followed the two-branch pipeline of faster-RCNN~\cite{ren2015faster}
and proposed a straight line detection framework to find the meaningful semantic straight line in an image. 
One branch verifies the existence of a line and the other branch further refines the position of the line
by regression.
Zhang \etal \cite{zhang2019ppgnet} adopted the conception of CornerNet~\cite{law2018cornernet} 
to extract line segments as a pair of key points in indoor scenes.
Huang \etal \cite{huang2018learning} proposed a two-head network to predict lines and junction points for wireframe parsing.
This is extended in ~\cite{zhou2019end} by adding a line proposal sub-network.
Zhou \etal \cite{zhou2019end} proposed an end-to-end architecture to perform accurate line segments detection in wireframe parsing. 

All these methods extract line-wise feature vectors by LoI pooling that aggregate deep features solely along each line,
leading to inadequate context information.

\subsection{Semantic Line Detection}
The meaningful straight line which helps photographic composition was firstly discussed in~\cite{lee2017semantic},
and named as ``semantic line''.
\cite{lee2017semantic} regarded semantic line detection as a special case of object detection.
It first extracts CNN representations of line proposals using LoI pooling, which bilinearly interpolates the 
features along the entire straight line.
Then the line representations are verified by a classifier and a regressor, similar to Faster-RCNN~\cite{girshick2015fast}.
The line proposals are all unique lines in an image.
The metric of the intersection of union (IoU) of two straight lines is proposed in~\cite{lee2017semantic} to evaluate the similarity of two straight lines in an image.
This metric may produce ambiguous definitions in some scenarios, as will be mentioned in~\cref{sec:metric}.
Besides, Lee \etal \cite{lee2017semantic} collected a semantic line detection dataset which
contains about 1,700 outdoor images, and most of them are natural landscape.

\section{Approach}\label{sec:dht}
In this section, we give the details of the proposed deep Hough transform for semantic line
detection.
Our proposed method mainly contains four components:
1) a CNN encoder that extracts pixel-wise deep representations;
2) the deep Hough transform (DHT) that converts the deep representations from the spatial domain to the parametric domain;
3) a line detector that is responsible for detecting lines in the parametric space, and
4) a reverse Hough transform (RHT) component that converts the detected lines back to image space.
All these components are integrated in an end-to-end framework that performs forward inference and backward training within a single step.
\revise{The pipeline is illustrated in~\cref{fig:pipeline}, and the detailed architecture is shown in the supplementary materials.}

\CheckRmv{
\begin{figure*}[tb!]
  \centering
  \begin{overpic}[scale=0.52]{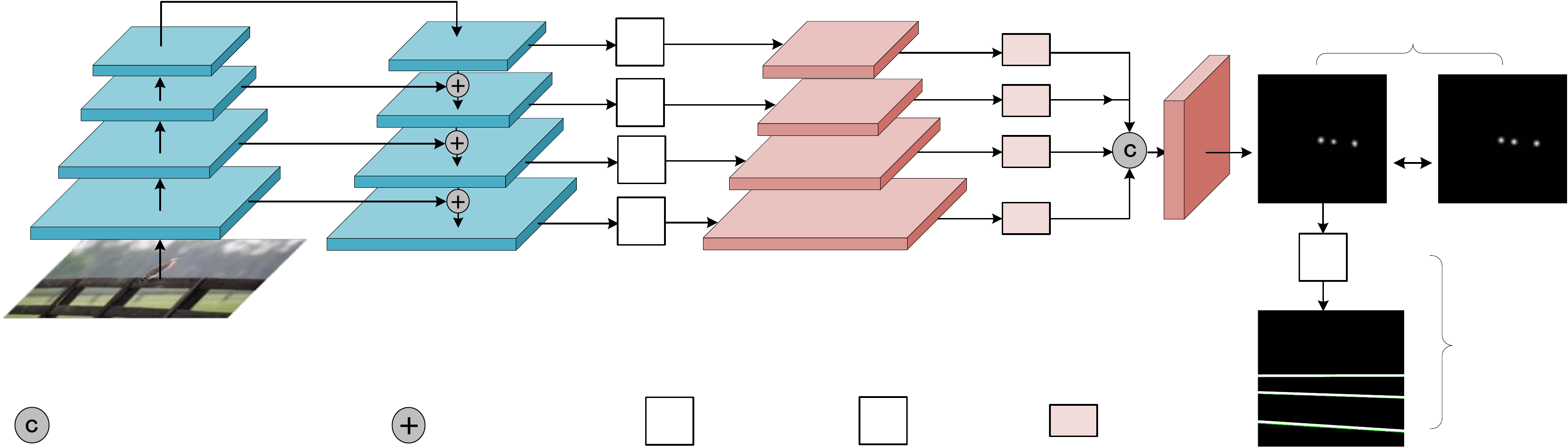}
    \put(31, 28){$X$}
    \put(54, 28){$Y$}
    \put(40.2, 13.9){$\mathcal{H}$}
    \put(40.2, 17.65){$\mathcal{H}$}
    \put(40.2, 21.45){$\mathcal{H}$}
    \put(40.2, 25.35){$\mathcal{H}$}
    \put(83.65, 11.5){$\mathcal{R}$}
    \put(41.7, 1){$\mathcal{H}$}
    \put(55.3, 1){$\mathcal{R}$}

    \put(67.15, 1.2){\scriptsize CTX}

    \put(64.1, 14.2){\scriptsize CTX}
    \put(64.1, 18.4){\scriptsize CTX}
    \put(64.1, 21.5){\scriptsize CTX}
    \put(64.1, 25){\scriptsize CTX}

    \put(88.6, 19){\scriptsize Loss}
    \put(85, 27){\small Training only}
    \put(93.5, 7.5){\small Testing}
    \put(94.5, 5){\small only}
    \put(5, 0.8){Upsample + Concat}
    \put(30, 0.8){Add}
    \put(47, 0.8){DHT}
    \put(60, 0.8){RHT}
    \put(72, 0.8){CTX}
  \end{overpic}
  \caption{
    \revise{The pipeline of our proposed method. DHT is short for the proposed Deep Hough Transform,
    and RHT represents the Reverse Hough Transform. CTX means the context-aware line detector which contains 
    multiple convolutional layers.}
  }\label{fig:pipeline}
\end{figure*}
}

\subsection{Line Parameterization and Reverse} \label{sec:preliminary}
\CheckRmv{
\begin{figure}[!htb]
  \centering
    \begin{overpic}[height=.6\linewidth]{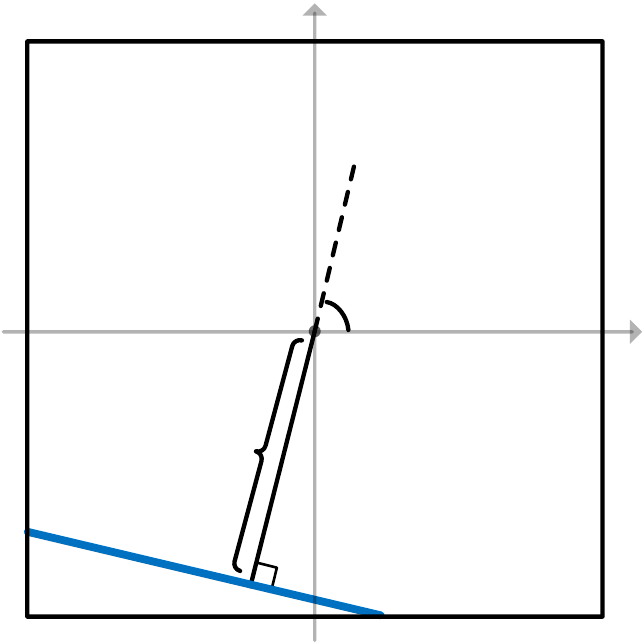}
      \put(29, 29){$r_l$}
      \put(57, 55){$\theta_l$}
      \put(100, 50){$x$}
      \put(52, 100){$y$}
    \end{overpic}
  \caption{A line can be parameterized by bias $r_l$ and slope $\theta_l$.
  }\label{fig:param}
\end{figure}
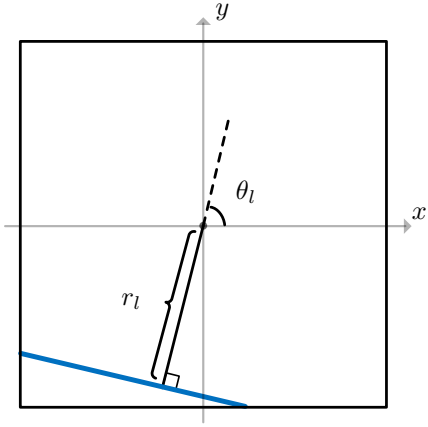
}
As shown in~\cref{fig:param},
given a 2D image \revise{$I_{H\times W}\in\mathbb{R}^{H\times W}$},
we set the origin to the center of the image.
In the 2D plane, a straight line $l$ can be parameterized by two parameters: an orientation parameter $\theta_l\in [0, \pi)$ representing the angle between $l$ and the x-axis and a distance parameter $r_l$, indicating the distance between $l$ and the origin.
Obviously $\forall \ l \in I, r_l \in [-\sqrt{W^2+H^2}/2, \sqrt{W^2+H^2}/2]$.

Given any line $l$ on $I$, we can parameterize it with the above formulations,
and also we can perform a reverse mapping to translate any valid
($r, \theta$) pair to a line instance.
\revise{We define the line-to-parameters and the inverse mapping
as:
\begin{equation}
\begin{split}
  r_l, \theta_l &\leftarrow P(l), \\
  l             &\leftarrow P^{-1}(r_l, \theta_l).
  \label{eq:parameterize}
\end{split}
\end{equation}
Obviously, both $P$ and $P^{-1}$ are bijective mappings.}
In practice, $r$ and $\theta$ are quantized to discrete bins to be processed by computer programs.
Suppose the quantization interval for $r$ and $\theta$ are $\Delta r$ and $\Delta \theta$, respectively.
Then the quantization can be formulated as below:
\begin{equation}
  \begin{split}
    \hat{r}_l = \ceil*{\frac{r_l}{\Delta r}}, \
    \hat{\theta}_l = \ceil*{\frac{\theta_l}{\Delta \theta}},
  \end{split}\label{eq:quantization}
\end{equation}
where $\hat{r}_l$ and $\hat{\theta_l}$ are the quantized line parameters.
The number of quantization levels, denoted with $\Theta$ and $R$, are:
\begin{equation}
  \begin{split}
    \Theta = \frac{\pi}{\Delta \theta}, \
    R = \frac{\sqrt{W^2+H^2}}{\Delta r},
  \end{split}\label{eq:grid-size}
\end{equation}
as shown in~\cref{fig:DHT}.

\subsection{Feature Transformation with Deep Hough Transform} \label{sec:dht-dht}

\CheckRmv{
\begin{figure*}[tb]
  \centering
  \hfill
  \subfigure[]{
    \label{fig:DHT}
    \begin{overpic}[height=.18\linewidth]{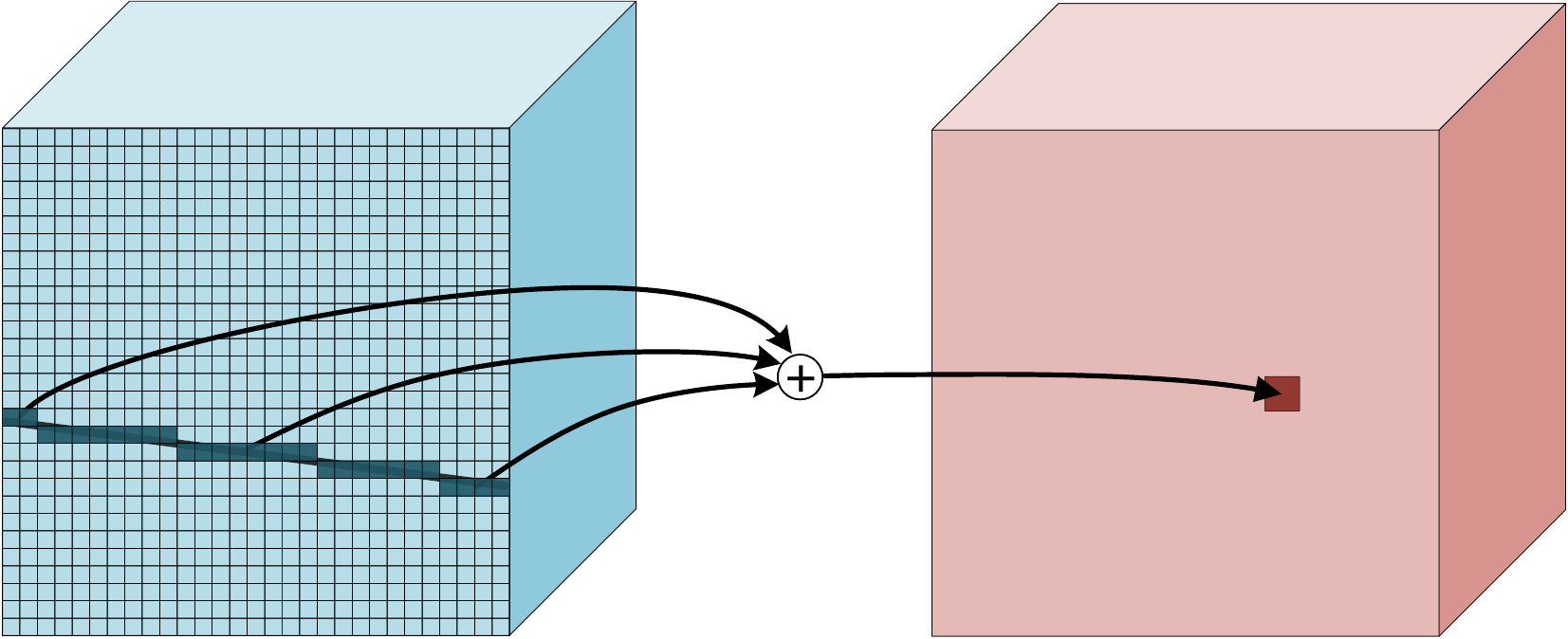}
      \put(-5, 19){$H$}
      \put(15, 33.3){$W$}
      %
      \put(55, 19){$\Theta$}
      \put(75, 33.3){$R$}
      \put(20, 43){$\mathbf{X}$}
      \put(79, 43){$\mathbf{Y}$}
      \put(76, 8){$(\hat{\theta}_l, \hat{r}_l)$}
    \end{overpic}
  }\hfill
  \subfigure[]{
    \label{fig:nonlocal}
    \begin{overpic}[width=0.46\linewidth]{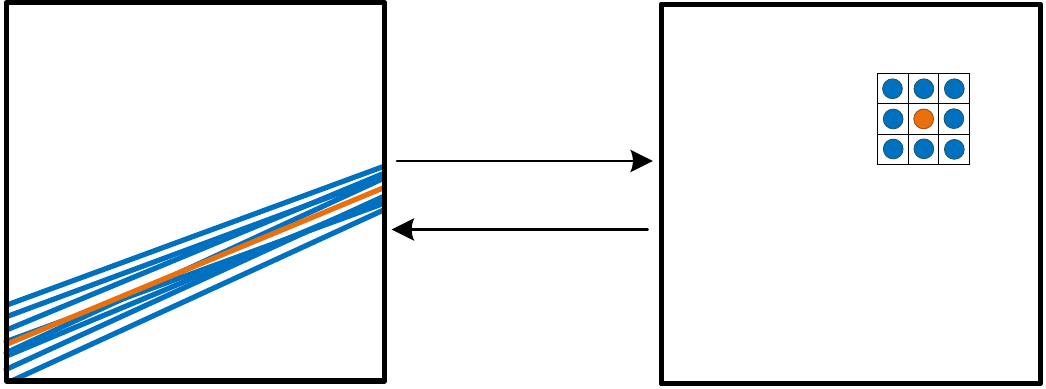}
      \put(19, 39){$W$}
      \put(-6, 20){$H$}
      \put(80, 39){$\Theta$}
      \put(102, 20){$R$}
      \put(45, 24){DHT}
      \put(45, 11){RHT}
      \put(6, -4){feature space}
      \put(65, -4){parametric space}
    \end{overpic}
  }\hfill
  \caption{
    (a): Features along a line in the feature space (blue, left) are
    accumulated to a point $(\hat{r}_l,\hat{\theta}_l)$ in the parametric space (red, right).
    (b): Illustration of the proposed context-aware feature aggregation.
    Features of nearby lines in the feature space (left)
    are translated into neighbor points in the parametric space (right).
    In the parametric space, a simple $3\times 3$ convolutional operation can
    easily capture contextual information for the central line (orange).
    Best viewed in color.
  }
\end{figure*}
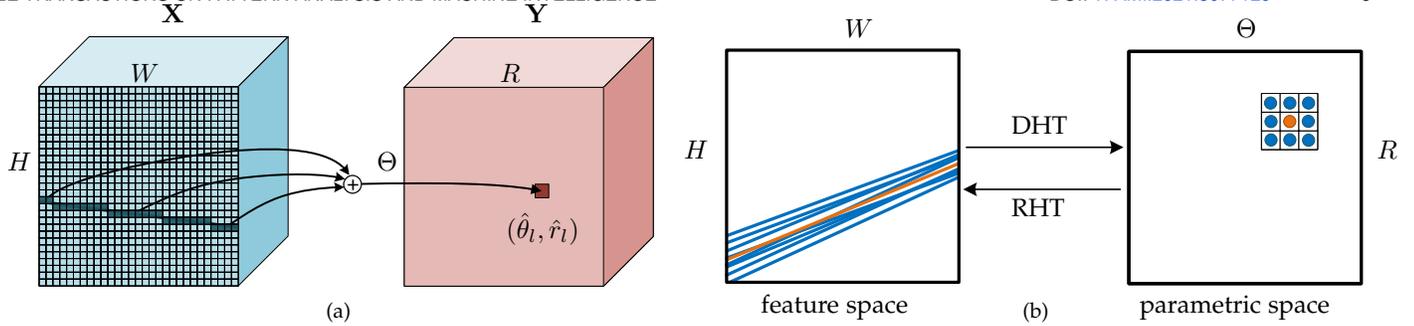
}

\subsubsection{Deep Hough transform.}
Given an input image $I$,  we first extract deep CNN features 
$\mathbf{X} \in \mathbb{R} ^ {C \times H \times W}$ with the encoder network,
where $C$ indicates the number of channels and $H$ and $W$ are the spatial size.
Afterward, the deep Hough transform (DHT) takes $\mathbf{X}$ as input and produces
the transformed features, $\mathbf{Y} \in \mathbb{R} ^ {C \times \Theta \times R}$.
The size of transformed features, $\Theta, R$, is determined by the quantization intervals, as
described in~\cref{eq:grid-size}.

As shown in~\cref{fig:DHT},
given an arbitrary line $l$ on the image,
we aggregate features of all pixels along $l$,
to $(\hat{\theta}_l, \hat{r}_l)$ in the parametric space $Y$:
\begin{equation}
  \mathbf{Y}(\hat{\theta}_l, \hat{r}_l) = \sum_{i\in l} \mathbf{X}(i),
  \label{eq:dht}
\end{equation}
where $i$ is the positional index.
$\hat{\theta}_l$ and $\hat{r}_l$ are determined by the parameters of line $l$, according
to~\cref{eq:parameterize}, and then quantized into discrete grids, according to~\cref{eq:quantization}.

\revise{Given the number of quantization levels $\Theta$ and $R$,
we have $\Theta\cdot R$ unique line candidates.
Then the DHT is applied to all these candidate lines and their respective features
are aggregated to the corresponding position in $\mathbf{Y}$.}
%
%
It is worth noting that DHT is order-agnostic in both the feature space and the parametric space,
making it highly parallelizable.


\subsubsection{Multi-scale DHT with FPN.}\label{sec:ms-dht-fpn}
Our proposed DHT could be easily applied to arbitrary spatial features.
We use the feature pyramid network (FPN)~\cite{lin2017feature} as our encoder.
FPN can help to extract multi-scale and rich semantic features.

Specifically, the FPN outputs 4 feature maps $X_1, X_2, X_3, X_4$ and their respective
resolutions are $1/4$, $1/8$, $1/16$, $1/16$ of the input resolution.
Then each feature map is transformed by a DHT module independently, as shown in~\cref{fig:pipeline}.
Since these feature maps are in different resolutions, the transformed features
$Y_1, Y_2, Y_3, Y_4$ also have different sizes, because we use the same quantization
interval in all stages (see~\cref{eq:grid-size} for details).
To fuse transformed features together, we interpolate $Y_2, Y_3, Y_4$
to the size of $Y_1$, and then fuse them by concatenation.

\subsection{Line Detection in the Parametric Space}


\subsubsection{Context-aware line detector.}\label{sec:ctx-line-detector}
After the deep Hough transform (DHT), features are translated to the parametric space
where grid location $(\theta, r)$ corresponds to
features along an entire line $l=P^{-1}(\theta, r)$ in the feature space.
An important reason to transform the features into the parametric space 
is that the line structures could
be more compactly represented.
As shown in~\cref{fig:nonlocal},
lines nearby a specific line $l$ are translated to
surrounding points near $(\theta_l, r_l)$.
Consequently, features of nearby lines can be efficiently aggregated
using convolutional layers in the parametric space.

In each stage of the FPN, 
we use two $3\times 3$ convolutional  layers
to aggregate contextual line features.
Then we interpolate features
to match the resolution of features from different stages, as illustrated in
~\cref{fig:pipeline},
and concatenate the interpolated features together.
Finally, a $1\times 1$ convolutional  layer is applied to the concatenated feature maps
to produce pointwise predictions.


\subsubsection{Loss function.}\label{sec:loss-func}
Since the prediction is directly produced in the parametric space,
we calculate the loss in the same space as well.
For a training image $I$, the ground-truth lines are first converted into
the parametric space with the standard Hough transform.
Then to help converging faster, we smooth and expand the ground-truth with a
Gaussian kernel.
Similar tricks have been used in many other tasks like
crowed counting~\cite{liu2019context,cheng2019learning} and road segmentation~\cite{VecRoad_20CVPR}.
Formally, let $\mathbf{G}$ be the binary ground-truth map in the parametric space,
$\mathbf{G}_{i,j} = 1$ indicates there is a line located at $i,j$ in the parametric space.
The expanded ground-truth map is
$$\hat{\mathbf{G}} = \mathbf{G}\circledast K,$$
where $K$ is a $5\times 5$ Gaussian kernel and $\circledast$ denotes the convolution operation.
An example pair of smoothed ground-truth and the predicted map is shown
in~\cref{fig:pipeline}.

In the end, we compute the cross-entropy between the smoothed ground-truth
and the predicted map in the parametric space:

\begin{equation}
  L = -\sum_i \Big\{ \hat{\mathbf{G}}_i\cdot\log(\mathbf{P}_i) +
                           (1-\hat{\mathbf{G}}_i)\cdot\log(1-\mathbf{P}_i)
                      \Big\}
\end{equation}

\subsection{Reverse Mapping}\label{sec:reverse}
Our detector produces predictions in the parametric space representing
the probability of the existence of lines.
The predicted map is then binarized with a threshold (\eg 0.01).
Then we find each connected area and calculate respective centroids.
These centroids are regarded as the parameters of detected lines.
At last, all lines are mapped back to the image space with
$P^{-1}(\cdot)$, as formulated in~\cref{eq:parameterize}.
We refer to the ``mapping back'' step as ``Reverse Mapping of Hough Transform (RHT)'',
as shown in~\cref{fig:pipeline}.

\subsection{Edge-guided Line Refinement}\label{sec:refine} 
Semantic lines are outstanding structures that
separate different regions in a scene.
Therefore, edges may serve as indicators for semantic lines.
We propose to refine the detection results by aligning line positions using
edge information.
First, we compute an edge map $E$ using HED~\cite{xie2015holistically}.
Afterward, given a detected line $l$, the edge density of $l$ is defined as the average
edge response along $l$:
\begin{equation}
\rho(l) = \frac{\sum_{i\in l} E_i}{|l|},
\label{eq:rho}
\end{equation}
where $|l|$ is the number of pixels on $l$.
For the sake of stability, we widen $l$ by 1 pixel on both sides (totally the width is 3)
when dealing with~\cref{eq:rho}.

Let $\mathcal{L}$ be a set of lines that are close to $l$.
These lines are obtained by moving the end-points of $l$ by $\delta_r$ pixels clockwise
and anti-clockwise.
Since there are two end-points and each one has
$\delta_r+1$ possible locations, the size of the set is $||\mathcal{L}|| = (\delta_r+1)^2$.
Then the refinement can be achieved by finding the optimal line $l^*$ from $\mathcal{L}$
that holds the highest edge density:
\begin{equation}
  l^* = \argmax_{l\in\mathcal{L}} \ \rho(l).\label{eq:refine-search}
\end{equation}
The performance of ``edge-guided line refinement'' with different $\delta_r$
is recorded in ~\cref{sec:ablation-refinement}.



\CheckRmv{
\begin{figure*}[tb!]
  \centering
  \subfigure[]{
    \begin{overpic}[width=.19\linewidth]{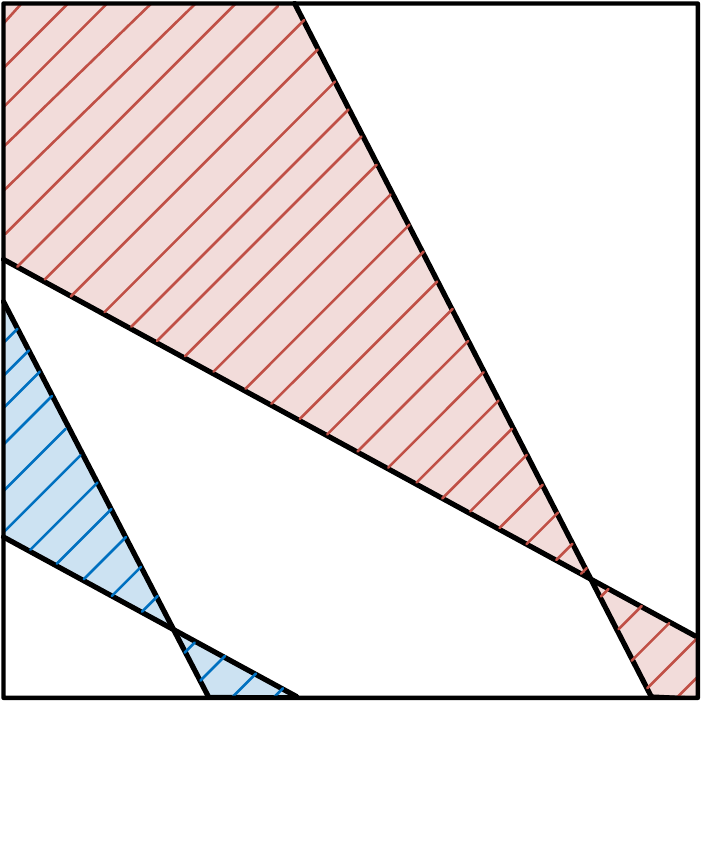}
      \put(54,66){m}
      \put(74,32){n}
      \put(10,50){p}
      \put(7,25){q}
      \put(19,10){{IOU(m,n)=0.66}}
      \put(20,0){{IOU(p,q)=0.95}}
    \end{overpic}\label{fig:metric_a}
    }
  \hfill
  \subfigure[]{
    \begin{overpic}[width=.19\linewidth]{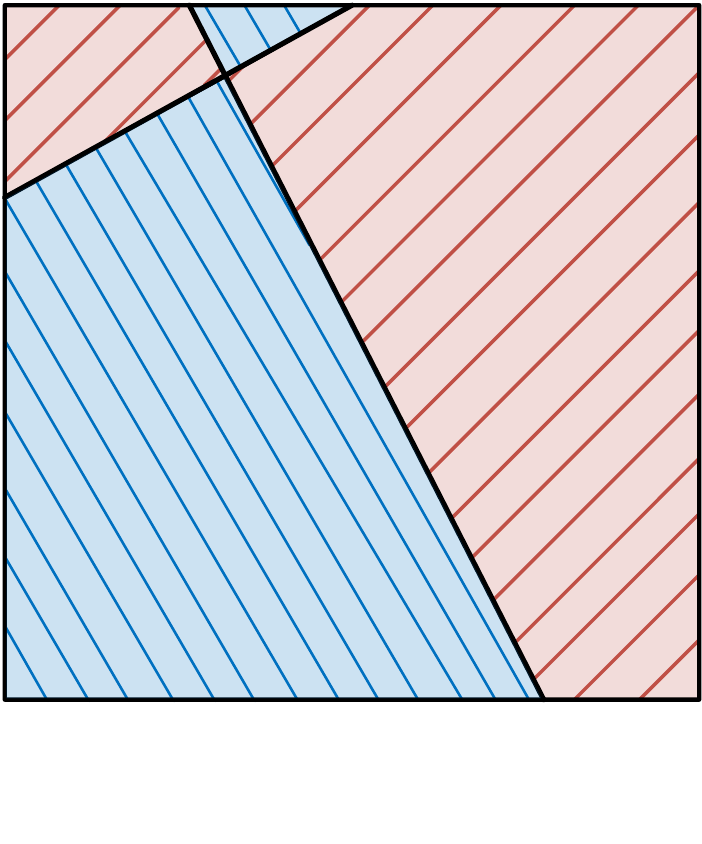}
      \put(18,10){{IOU(red)=0.53}}
      \put(16.5,0){{IOU(blue)=0.46}}
    \end{overpic}\label{fig:metric_b}
  }
  \hfill
  \subfigure[]{
    \begin{overpic}[width=.19\linewidth]{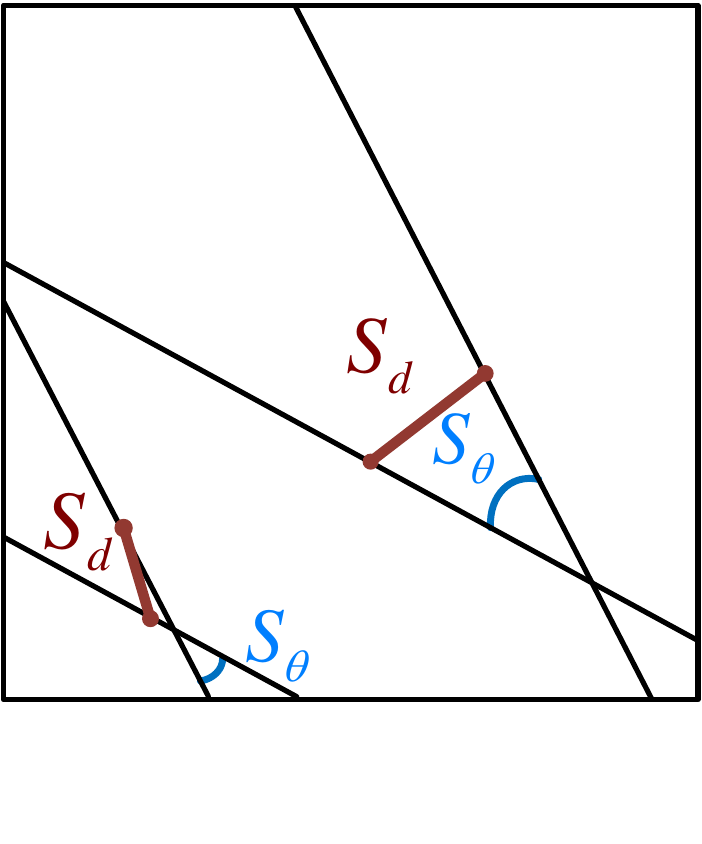}
      \put(54,66){m}
      \put(74,32){n}
      \put(10,50){p}
      \put(7,25){q}
      \put(22,10){{$\mathcal{S}$(m,n)=0.24}}
      \put(23,0){{$\mathcal{S}$(p,q)=0.34}}
    \end{overpic}\label{fig:metric_c}
  }
  \hfill
  \subfigure[]{
    \begin{overpic}[width=.19\linewidth]{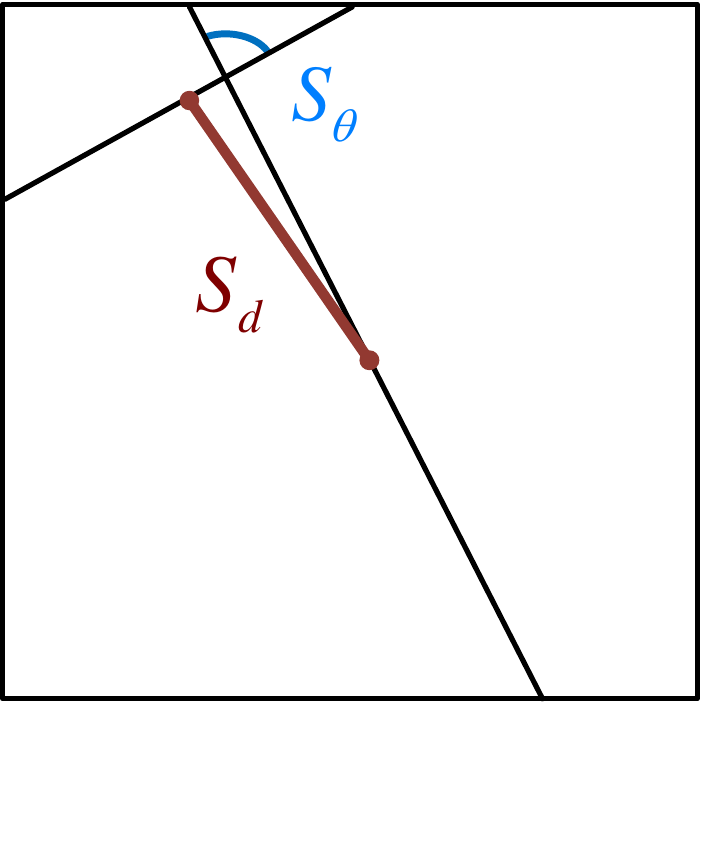}
      \put(16,10){{$\mathcal{S}_d$=0.54, $\mathcal{S}_\theta$=0.02}}
      \put(28.5,0){{$\mathcal{S}$=0.00}}
    \end{overpic}\label{fig:metric_d}
  }\vspace{-10pt}
  \caption{
    (a): Two pairs of lines with similar relative position could have very
        different IOU scores.
    (b): Even humans cannot determine which area (blue or red) should be considered as
    the intersection in the IOU-based metric~\cite{lee2017semantic}.
    (c) and (d): Our proposed metric considers both Euclidean distance and angular distance
        between a pair of lines, resulting in consistent and reasonable scores.
    Best viewed in color.
  }
  \label{fig:metric_analysis}
\end{figure*}
}

\CheckRmv{
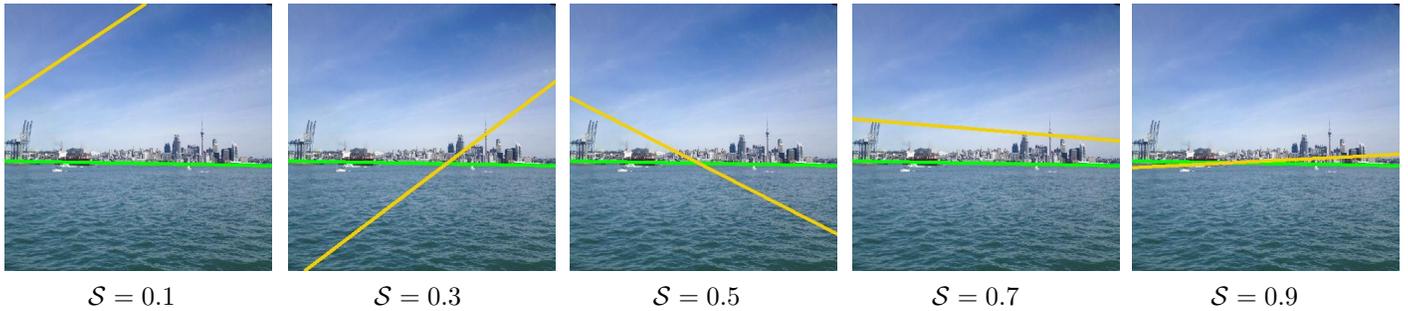
\begin{figure*}[t!]
  \centering
  \begin{overpic}[width=1.0\linewidth]{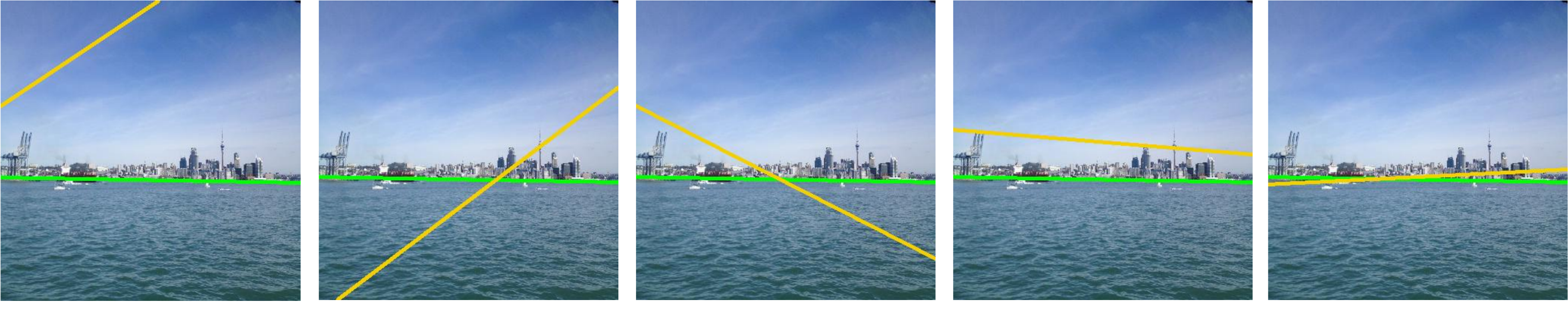}
    \put(6, -1.5){$\mathcal{S}=0.1$}
    \put(26.5, -1.5){$\mathcal{S}=0.3$}
    \put(46.5, -1.5){$\mathcal{S}=0.5$}
    \put(66.5, -1.5){$\mathcal{S}=0.7$}
    \put(86.5, -1.5){$\mathcal{S}=0.9$}
\end{overpic}
  \caption{
    Example lines with various EA-scores ($\mathcal{S}$ in ~\cref{eq:ea-score}).
    \revise{The larger the EA-score is, the more similar the lines are.}
  }\label{fig:metric_show}
\end{figure*}
}

\section{The Proposed Evaluation Metric} 
\label{sec:metric}

In this section, we elaborate on the 
proposed evaluation metric that measures the agreement, or alternatively, 
the similarity between the two lines in an image.
Firstly, we review several widely used metrics in the computer vision community 
and then explain why these existing metrics are not proper for our task.
Finally, we introduce our newly proposed metric, which measures the agreement between two lines considering
both Euclidean distance and angular distance.
%

\subsection{Review of Existing Metrics}
The intersection over union (IOU) is widely used in object detection, semantic segmentation
and many other tasks to measure the agreement between detected bounding boxes
(segments) w.r.t the ground-truth.
Lee \etal \cite{lee2017semantic} adopt the original IOU into line detection,
and propose the line-based IOU to evaluate the quality of detected lines.
Concretely, the similarity between the two lines is measured by the intersection areas
of lines divided by the image area.
Take~\cref{fig:metric_a} as an example, the similarity between line $m$ and $n$
is $\text{IOU}(m,n) = area({{red}})/area(I)$.

However, we argue that this IOU-based metric is improper  and may lead to unreasonable
or ambiguous results under specific circumstances.
As illustrated in~\cref{fig:metric_a}, 
two pairs of lines (m, n, and p, q) with similar structures could have
very different IOU scores.
In~\cref{fig:metric_b}, even humans cannot determine which areas
({red} or {blue}) should be used as
intersection areas in line based IOU.

\rerevise{There are other metrics,
\eg the Earth Mover's Distance (EMD)~\cite{rubner2000earth}
and the Chamfer distance (CD)~\cite{borgefors1986distance},
that can be used to measure line similarities.
However, these metrics require to rasterize the lines into pixels
and then calculate pixel-wise distances, which is less efficient.
}

\rerevise{
To remedy the deficiencies,
we propose a simple yet effective metric that measures the similarity 
of two lines in the parametric space.
Our proposed metric is much more efficient than EMD and CD.
Quantitative comparisons in ~\cref{sec:quantitative} demonstrate that 
our proposed metric presents very similar results to EMD and CD.}

\CheckRmv{
\begin{figure*}[!htb]
  \centering
  \begin{overpic}[width=1\linewidth]{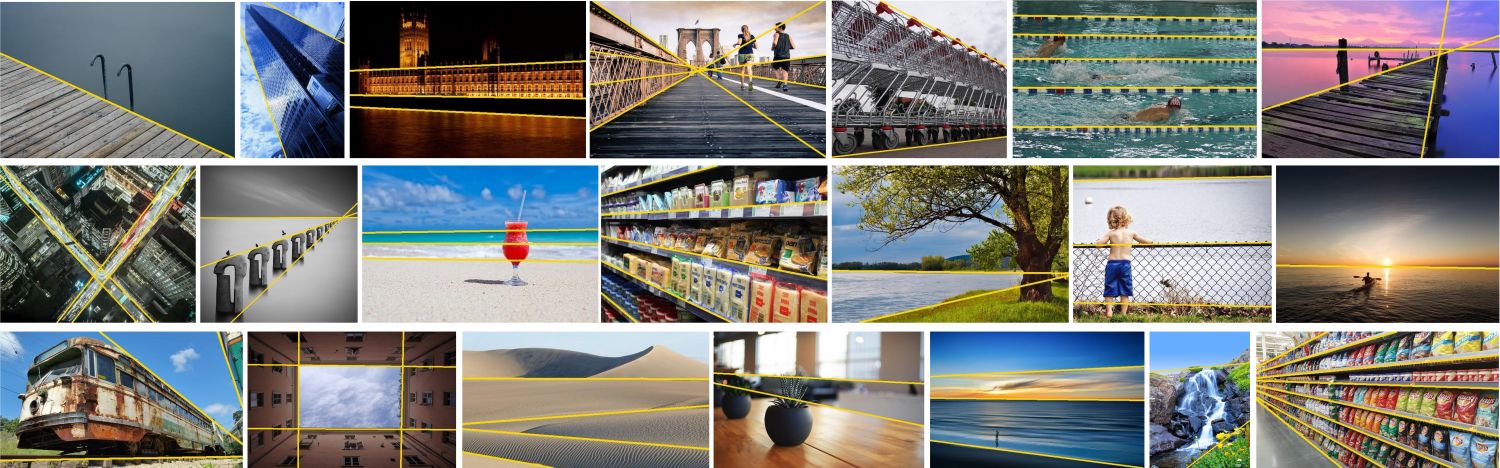}
  \end{overpic}\vspace{-8pt}
  \caption{
    Example images and annotations (yellow lines) of \revise{NKL}.
    Images of \revise{NKL} present diverse scenes and rich line annotations.
  }
  \label{fig:dataset_vis}
\end{figure*}
}

\subsection{The Proposed Metric}
Our proposed metric, \revise{termed \textbf{EA-score}}, considers both 
\textbf{E}uclidean distance and \textbf{A}ngular distance
between a pair of lines.
Let $l_i, l_j$ be a pair of lines to be measured,
the angular distance $\mathcal{S}_\theta$ is defined according to the
angle between two lines:
\begin{equation}
    \mathcal{S}_\theta = 1 - \frac{\theta(l_i, l_j)}{\pi/2},
    \label{eq:sa}
\end{equation}
where $\theta(l_i, l_j)$ is the angle between $l_i$ and $l_j$.
The Euclidean distance is defined as:
\begin{equation}
    \mathcal{S}_d = 1 - D(l_i, l_j),
    \label{eq:sd}
\end{equation}
where $D(l_i, l_j)$ is the Euclidean distance between midpoints of $l_i$ and $l_j$.
%
%
Note that we normalize the image into a unit square before calculating $D(l_i, l_j)$.
Examples of $\mathcal{S}_d$ and $\mathcal{S}_\theta$ can be found in
~\cref{fig:metric_c} and~\cref{fig:metric_d}.
Finally, our proposed EA-score is:
\begin{equation}
    \mathcal{S} = (\mathcal{S}_\theta \cdot \mathcal{S}_d)^2.
    \label{eq:ea-score}
\end{equation}
~\cref{eq:ea-score} is squared to make it more sensitive and discriminative
when the values are high.

Several example line pairs and corresponding EA-scores
are demonstrated in~\cref{fig:metric_show}.

\section{\revise{NKL: a semantic line detection dataset}}\label{sec:nkl-dataset}
To the best of our knowledge, there is only one dataset, SEL~\cite{lee2017semantic},
specifically for semantic line detection.
SEL contains 1,715 images of which 175 images for testing
and others for training.
To fulfill the gap between large CNN-based models and \revise{the} scale of the existing
dataset,
we collect a new dataset for semantic line detection.

The new dataset, \revise{namely NKL (short for \textbf{N}an\textbf{K}ai \textbf{L}ines)
contains 6,500 images} that present richer
diversity in terms of both scenes and the number of lines.
Each image of \revise{NKL} is annotated by multiple skilled human annotators to
ensure the annotation quality.

\CheckRmv{
\begin{table}[hbt!]
  \renewcommand{\arraystretch}{1.3}
  \renewcommand\tabcolsep{5pt}
  \caption{Number of images and lines in SEL~\cite{lee2017semantic} and \revise{NKL}.}
  \vspace{-5pt}
  \begin{tabular}{c|c|c|c}
    \toprule  
    Dataset &
    \makecell{Total \\ \#images, \#lines} &
    \makecell{Training \\ \#images, \#lines} &
    \makecell{Evaluation \\ \#images, \#lines} \\
    \hline
    SEL~\cite{lee2017semantic} & 1,715, \ 2,791 & 1,541, \ 2,493 & 175, \ 298\\
    \revise{NKL (Ours)} & \revise{6,500}, \ \revise{13,148} & \revise{5,200}, \ \revise{10,498} & \revise{1,300, \ 2,650} \\
    \bottomrule
  \end{tabular}
  \label{tab:number_statistics}
\end{table}
}

\subsection{Data Collection and Annotation}\label{sec:data-collect-and-anno}
All the images of \revise{NKL} are crawled from the internet
using specific keywords such as sea, grassland \etal
After copyright checking, we carefully filter out images with
at least one semantic line.
Since the annotation of semantic lines is subjective and depends on annotators,
each image is first annotated by 3 knowledgeable human annotators
and verified by others.
A line is regarded as positive only if all of the 3 annotators are consistent.
Then the inconsistent lines are reviewed by two other annotators.
In a word, for each line, there are at least 3 and at most 5 annotators,
and a line is regarded as positive only if the line is marked as positive by more than 3 annotators.

\subsection{Dataset Statistics}
\subsubsection{Number of images and semantic lines}
%
%
There are totally \revise{13,148 semantic lines in NKL} and 2,791 semantic lines in
SEL~\cite{lee2017semantic} dataset over all images.
\cref{tab:number_statistics} summarizes the number of images and lines of the two datasets, respectively.

~\cref{fig:number_statistics} summarizes the histogram of the per-image number of lines
in \revise{NKL} and SEL~\cite{lee2017semantic} datasets.
More than half \revise{(67\%, 4,356/6,500)} of the images in \revise{NKL} dataset contain more than 1 semantic line,
while the percentage of SEL is only 45.5\%.

\CheckRmv{
\begin{figure}[!htb]
  \centering
  \begin{overpic}[width=1\linewidth]{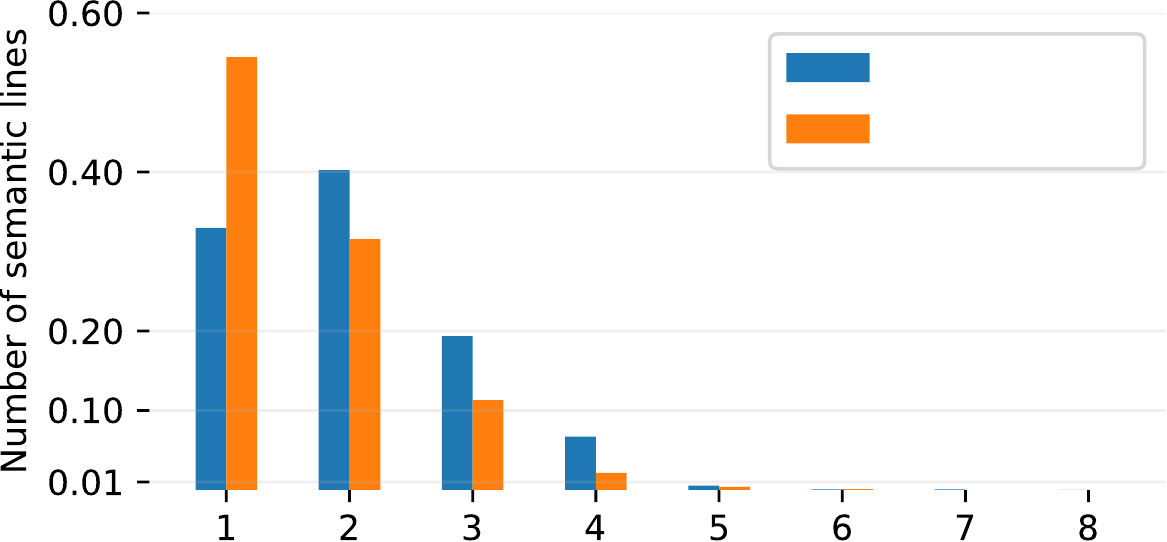}
    \put(15, 26){\rotatebox{30}{\scriptsize 33.0\%}}
    \put(19, 41){\rotatebox{30}{\scriptsize 54.5\%}}
    
    \put(26, 31){\rotatebox{29}{\scriptsize 40.2\%}}
    \put(30, 25.5){\rotatebox{30}{\scriptsize 31.5\%}}

    \put(37, 17.3){\rotatebox{30}{\scriptsize 19.4\%}}
    \put(40, 11.5){\rotatebox{30}{\scriptsize 11.3\%}}

    \put(47, 8.6){\rotatebox{30}{\scriptsize 6.7\%}}
    \put(51, 5){\rotatebox{30}{\scriptsize 2.1\%}}

    \put(58.5, 5.5){\rotatebox{30}{\scriptsize 0.54\%}}
    \put(62.2, 4.5){\rotatebox{30}{\scriptsize 0.41\%}}

    \put(69, 4.8){\rotatebox{30}{\scriptsize 0.01\%}}
    \put(73, 4.2){\rotatebox{30}{\scriptsize 0.01\%}}

    \put(80, 5){\rotatebox{30}{\scriptsize 0.01\%}}
    \put(83, 4){\rotatebox{30}{\scriptsize 0}}

    \put(90.5, 5){\rotatebox{30}{\scriptsize 0.003\%}}
    \put(93.5, 4){\rotatebox{30}{\scriptsize 0}}

    \put(78, 39.6){\scriptsize NKL (Ours)}
    \put(78, 34.4){\scriptsize SEL~\cite{lee2017semantic}}
  \end{overpic}
  \caption{
    \revise{Histogram chart of number of lines.
    Lines of our dataset are more fairly distributed compared to
    SEL.}
  }
  \label{fig:number_statistics}
\end{figure}
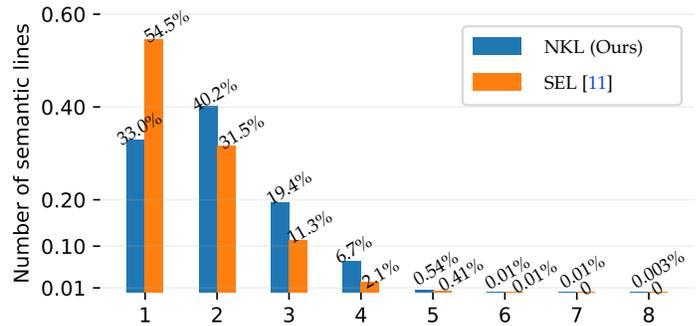
}

\subsubsection{Diversity Analysis}
To analyze the diversity of SEL and \revise{NKL} datasets,
we feed all the images into a ResNet50~\cite{he2016deep} network that is pretrained on the Place365~\cite{zhou2017places},
and then collect the outputs as category labels.
The results are presented in ~\cref{fig:dataset_distribution}.

There are totally 365 categories in Place365~\cite{zhou2017places} dataset,
among which we got 167 unique category labels on SEL dataset and \revise{327 on NKL}.
Besides, as shown in ~\cref{fig:dataset_distribution}, scene labels on
\revise{NKL} dataset are more fairly distributed compared to SEL dataset.
For example, in SEL dataset, top-3 populated categories (sky, field, desert) make up more
than a quarter of the total.
While in \revise{NKL}, top-3 makes up less than one-fifth of the total.



\CheckRmv{
\begin{figure*}[tb]
  \centering
  \hfill
  \subfigure[]{
    \begin{overpic}[width=.48\linewidth]{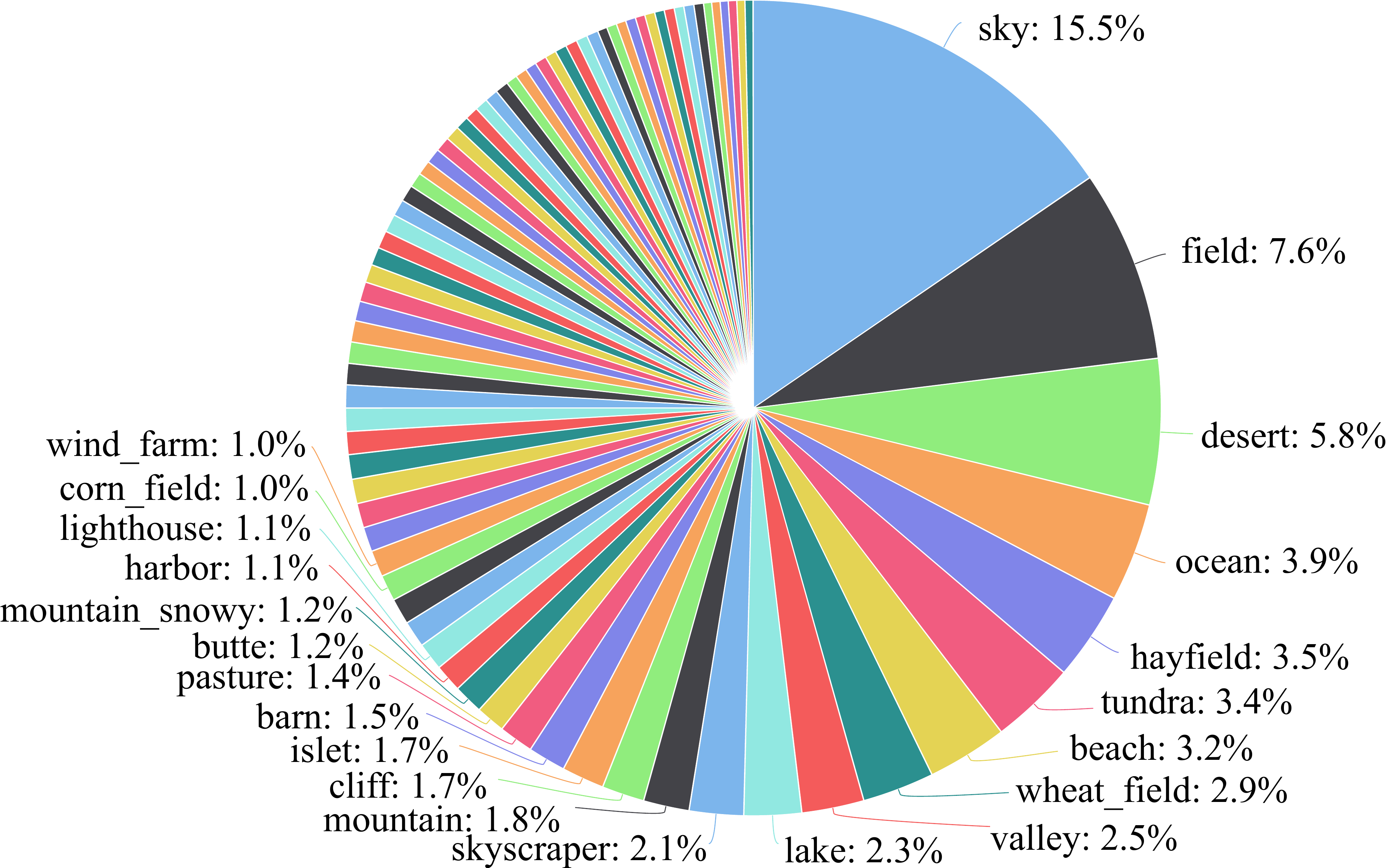}
      \put(6, 55){SEL~\cite{lee2017semantic}}
    \end{overpic}
  }\hfill
  \subfigure[]{
    \begin{overpic}[width=0.465\linewidth]{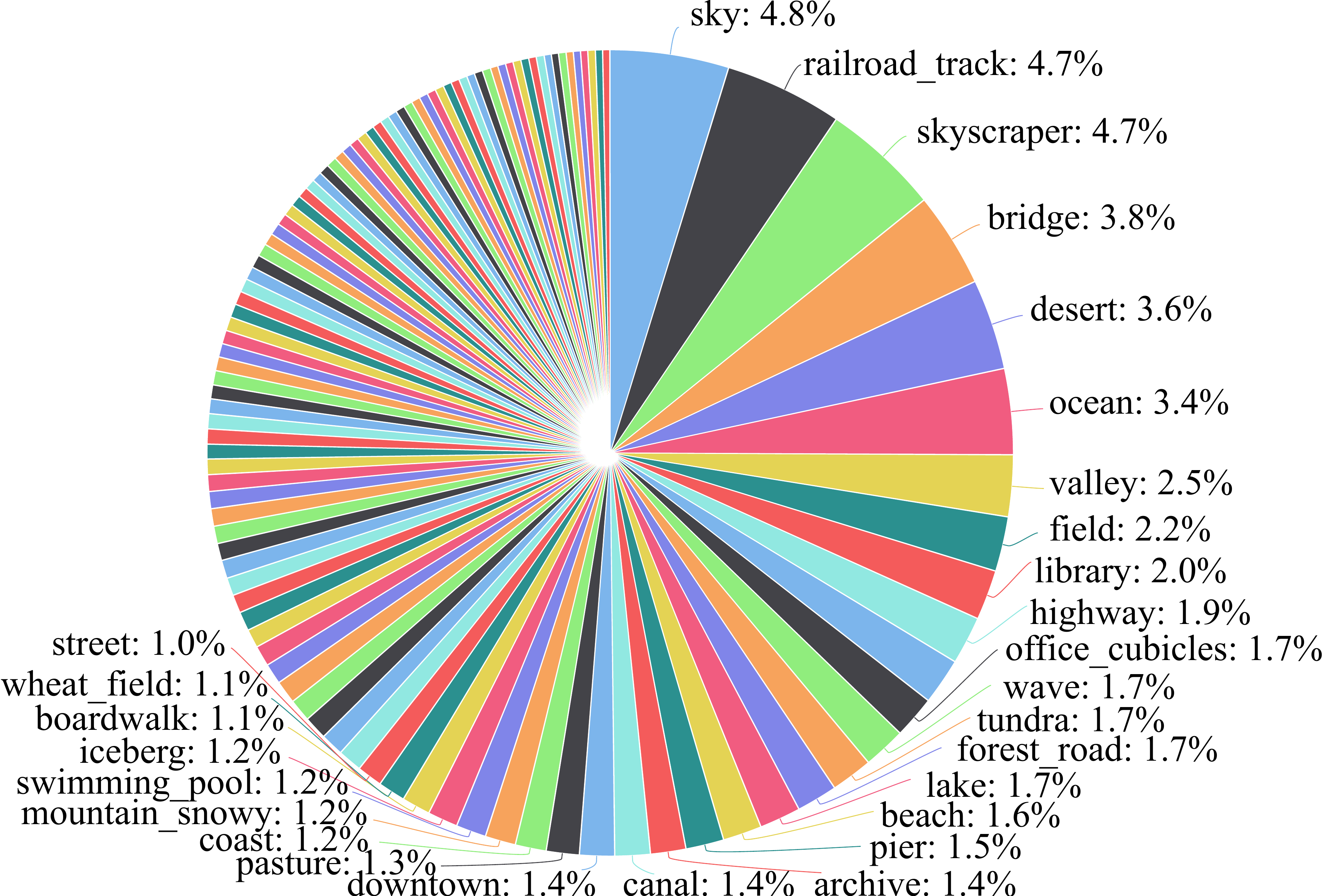}
      \put(5, 57){NKL}
    \end{overpic}
  }\hfill
  \vspace{-10pt}
  \caption{
    \revise{
    Category distribution of SEL (a) and NKL (b) datasets.
    Category labels are obtained through a Places365 pretrained model.
    There are 327 (totally 365) scene labels presented in NKL dataset,
    in contrast to 167 in SEL dataset.
    The labels of NKL are also more fairly distributed compared to
    that of SEL.}
  }
  \label{fig:dataset_distribution}
\end{figure*}
}

\section{Experiments}\label{sec:experiments}
In this section, we introduce the implementation details of our system,
and report experimental results compared with existing methods.

\subsection{Implementation Details} \label{sec:protocol}
Our system is implemented with the PyTorch~\cite{paszke2019pytorch} framework,
and a Jittor~\cite{hu2020jittor} implementation is also available.
Since the proposed deep Hough transform (DHT) is highly parallelizable,
we implement DHT with native CUDA programming,
and all other parts are implemented based on framework level Python API.
We use a single RTX 2080 Ti GPU for all experiments.

\subsubsection{Network architectures.}
We use two representative network architectures,
ResNet50~\cite{he2016deep} and VGGNet16~\cite{simonyan2014very}, as our backbone
and the FPN~\cite{lin2017feature} to extract multi-scale deep representations.
For the ResNet network, following the common practice in previous works~\cite{zhao2017pyramid,chen2018encoder,pami20Res2net},
the dilated convolution~\cite{yu2015multi} is used in the last layer to increase the resolution of feature maps. The common used batch normalization~\cite{ioffe2015batch} is also adopted in the network. 
The dilated convolutions and normalization can be tuned by~\cite{gao2021rbn,gao2021global2local} in the future work.


\subsubsection{Hyper-parameters.}
The size of the Gaussian kernel used in~\cref{sec:loss-func} is $5\times5$. 
All images are resized to (400, 400) and then wrapped into a mini-batch of 8.
We train all models for 30 epochs using Adam optimizer~\cite{kingma2014adam} without weight decay.
The learning rate and momentum are set to $2 \times 10^{-4}$ and 0.9, respectively.
The quantization intervals $\Delta\theta, \Delta r$ will be detailed in
~\cref{sec:quant-intervals} and~\cref{eq:intervals}.


\CheckRmv{
\begin{figure*}[!htb]
  \centering
  \begin{overpic}[width=1\linewidth]{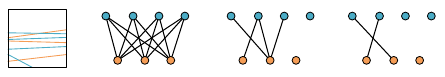}
    \put(-0.5,10){{\color{BlueGreen}{$p_1$}}}
    \put(-0.5,8){{\color{BlueGreen}{$p_2$}}}
    \put(-0.5,6){{\color{BlueGreen}{$p_3$}}}
    \put(-0.5,4.2){{\color{BlueGreen}{$p_4$}}}
    \put(15.5,11){{\color{orange}{$g_1$}}}
    \put(15.5,7.2){{\color{orange}{$g_2$}}}
    \put(15.5,4.5){{\color{orange}{$g_3$}}}
    \put(26,1.5){$g_1$}
    \put(31.5,1.5){$g_2$}
    \put(37.8,1.5){$g_3$}
    \put(22.5,15.2){$p_1$}
    \put(29,15.2){$p_2$}
    \put(35,15.2){$p_3$}
    \put(41.3,15.2){$p_4$}
    \put(54,1.5){$g_1$}
    \put(61,1.5){$g_2$}
    \put(67,1.5){$g_3$}
    \put(51.5,15.2){$p_1$}
    \put(57.3,15.2){$p_2$}
    \put(63.3,15.2){$p_3$}
    \put(70,15.2){$p_4$}
    \put(95.1,1.5){\scriptsize FN}
    \put(79,15.2){\scriptsize TP}
    \put(86,15.2){\scriptsize TP}
    \put(92,15.2){\scriptsize FP}
    \put(98,15.2){\scriptsize FP}
    \put(20,8){$\rightarrow$}
    \put(47,8){$\rightarrow$}
    \put(75,8){$\rightarrow$}
    \put(8,-1){(a)}
    \put(32,-1){(b)}
    \put(60,-1){(c)}
    \put(87.6,-1){(d)}
  \end{overpic}
  \caption{
    Illustration of the bipartite graph matching in evaluation.
    (a) An example image with 3 ground-truth lines ($g_1, g_2, g_3$)
    and 4 predictions ($p_1, p_2, p_3, p_4$).
    (b) the corresponding bipartite graph.
    The edge between a pair of nodes represents the similarity ($\mathcal{S}$ in ~\cref{eq:ea-score}) between lines.
    (c) after maximum matching of a bipartite graph, each node in a subgraph
    is connected with no more than 1 node from the other subgraph.
    (d) true positive (TP), false positive (FP) and false negative (FN).
  }\vspace{20pt}
  \label{fig:matching}
\end{figure*}
}


\CheckRmv{
\begin{figure*}[htb!]
  \centering
  \begin{overpic}[width=0.9\linewidth]{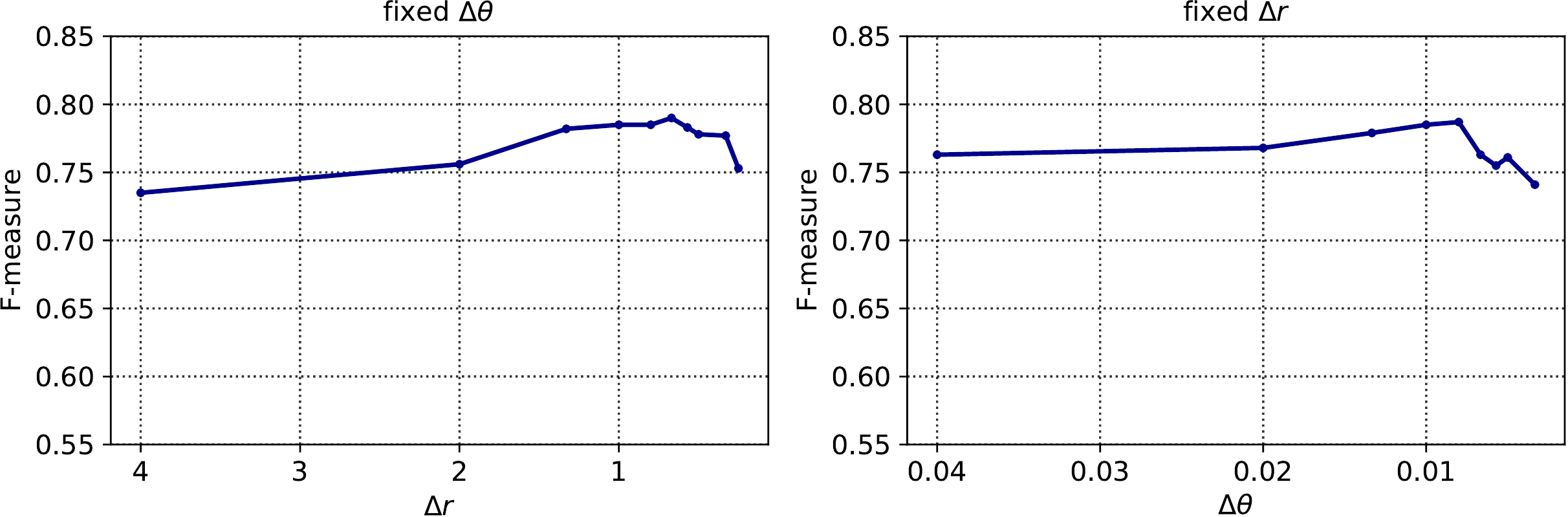}
    \put(95,1){$\times\pi$}
    \put(45,1){$\times\sqrt{2}$}
  \end{overpic}\vspace{-10pt}
  \caption{
    Left: performance under different distance quantization intervals $\Delta r$ with a fixed
         angular quantization interval $\Delta\theta=\pi/100$.
         Larger $\Delta r$ indicates less quantization levels $R$.
    Right: performance under different angular quantization intervals $\Delta \theta$ with a fixed
        distance quantization interval $\Delta r=\sqrt{2}$.
  }\label{fig:ablation-interval}
\end{figure*}
}

\subsubsection{Datasets and data augmentation.}
\revise{Our experiments are conducted on the 
SEL~\cite{lee2017semantic} dataset and our Proposed
NKL dataset.}
\revise{The Statistics of the two datasets are detailed in~\cref{sec:nkl-dataset}.}
%
%
%
Following the setup in~\cite{lee2017semantic}, we use only left-right flip
data augmentation in all our experiments.

\subsection{Evaluation Protocol}
We measure the quality of detection lines in terms of \emph{precision},
\emph{recall} and \emph{F-measure}.
The first step is to match the detected lines and ground-truth lines.

Let $\mathcal{P}$ and $\mathcal{G}$ be the sets of predicted lines
and ground-truth lines, respectively.
$p_i$ and $g_j$ are individual predicted and ground-truth line.
We first match the lines in $\mathcal{P}$ and $\mathcal{G}$ based on bipartite matching.
Suppose $G = \{V, E\}$ be a bipartite graph
\footnote{\url{https://en.wikipedia.org/wiki/Bipartite_graph}}.
The vertice set $V$ can be divided into two disjoint and independent sets,
in our case, $\mathcal{P}$ and $\mathcal{G}$:
\begin{equation*}
  \begin{split}
    V = \mathcal{P} \cup \mathcal{G} \\    
    \mathcal{P} \cap \mathcal{G} = \emptyset.
  \end{split}
\end{equation*}

\rerevise{
Each edge in $E$ denotes the similarity between a pair of lines under 
a certain similarity measure.
Apart from the proposed EA-score, we also use two other popular metrics:
the earth mover's distance (EMD)~\cite{rubner2000earth} and the Chamfer distance (CD)~\cite{borgefors1986distance}, as described in~\cref{sec:metric}.
Note that we normalize both EMD and chamfer distance by their maximal possible
value to bound the value
within [0, 1] (for both EMD and Chamfer distance, the maximal distance occurs when
two lines shrinkage to two points on the opposite diagonals).}

Given the graph $G = \{V, E\}$, a matching in a Bipartite Graph is a set of the edges chosen in such a way that no two edges share
a common vertice.
In our task, given the set of predicted lines $\mathcal{P}$ and the set of
ground-truth lines $\mathcal{G}$, we seek to find a matching so that
each ground-truth line $g_i$ corresponds to no more than one detected line $p_j$
and vice versa.
This problem, maximum matching of a bipartite graph, can be easily solved using the
classical Hungarian method~\cite{kuhn1955hungarian} with polynomial time complexity.

After matching $\mathcal{P}$ and $\mathcal{G}$, we can calculate true positive
(TP), false positive (FP) and false negative (FN) accordingly.
As illustrated in ~\cref{fig:matching}, predicted lines ($p_1, p_2$) that are paired with
ground-truth lines ($g_2, g_1$) are considered as true positive.
Predicted line ($p_3$) that is not matched with any ground-truth line is
a false positive,
and ground-truth line ($g_3$) without a corresponding predicted line
is a false negative.

Finally, the \textbf{P}recision, \textbf{R}ecall, and \textbf{F}-measure are:
\begin{equation}
    P = \frac{TP}{TP+FP}, \
    R = \frac{TP}{TP+FN}, \
    F = \frac{2PR}{P + R}.
    \label{eq:pr}
\end{equation}
We apply a series thresholds $\tau = 0.01, 0.02,...,0.99$ to prediction \& ground-truth pairs.
Accordingly, we derive a series of precision, recall, and F-measure scores.
Finally, we evaluate the performance in terms of average precision, recall, and
F-measure. \rerevise{We use EMD~\cite{rubner2000earth}, CD~\cite{borgefors1986distance}, and our proposed EA metric
for quantitative comparisons .
In the ablation study, we only use EA metric for simplicity.}

\subsection{\revise{Tuning the Quantization Intervals}}\label{sec:quant-intervals}
The quantization intervals $\Delta\theta$ and $\Delta r$ in~\cref{eq:quantization}
are important factors to the performance and running efficiency.
Larger intervals lead to fewer quantization levels, \ie $\Theta$ and $R$,
and the model will be faster.
With smaller intervals, there will be more quantization levels,
and the computational overhead is heavier.

\revise{We perform a coordinate descent on SEL~\cite{lee2017semantic} dataset
to find proper intervals that are computationally efficient and functionally
effective.
Note that we use the EA-score as line similarity measure since its simplicity.
In the first round, we fix the angular quantization interval to $\Delta\theta=\pi/100$
and then search for $\Delta r$,
the results are shown in ~\cref{fig:ablation-interval}(a).
According to ~\cref{fig:ablation-interval}(a),
the performance first rises slowly and then drops down with the decrease of $\Delta r$,
and the turning point is near $\Delta r = \sqrt{2}$.
In the second round, we fix $\Delta r = \sqrt{2}$ and train with different $\Delta\theta$.
Similar to ~\cref{fig:ablation-interval}(a),
the results in~\cref{fig:ablation-interval}(b) demonstrate that
the performance first increases smoothly with the drop of $\Delta \theta$,
and then quickly decreases with vibration.
Therefore, the turning point $\Delta\theta=\pi/100$ is a proper choice for angular quantization.}

In summary, we use $\Delta\theta=\pi/100$ and $\Delta r = \sqrt{2}$
in quantization, and corresponding quantization levels are:
\begin{equation}
    \Theta = 100, \ R = \sqrt{\frac{W^2+H^2}{2}},
  \label{eq:intervals}
\end{equation}
where $H, W$ are the size of feature maps to be transformed in DHT.
\subsection{Quantitative Comparisons} \label{sec:quantitative}
We compare our proposed method with the SLNet~\cite{lee2017semantic}
and the classical Hough line detection~\cite{duda1971use} with HED~\cite{xie2015holistically}
as the edge detector.
Note that we train the HED edge detector on the SEL~\cite{lee2017semantic} training set
using the line annotations as edge ground-truth.

\subsubsection{Results on SEL dataset}
\cref{tab:quantitative} summarizes the results on the SEL dataset~\cite{lee2017semantic}.
With either VGG16 or ResNet50 as the backbone, Our proposed method consistently outperforms
SLNet and HT+HED with a considerable margin.
In addition to~\cref{tab:quantitative}, we plot the F-measure \emph{v.s.} threshold and
the precision~\emph{v.s.} recall curves.
\cref{fig:curves} reveals that our method achieves higher F-measure than others
under a wide range of thresholds.

\CheckRmv{
\begin{table*}[!htb]
    \renewcommand{\arraystretch}{1.3}
    \renewcommand\tabcolsep{8.0pt}
    \centering
    \caption{
      Quantitative comparisons 
      on the SEL~\cite{lee2017semantic} and NKL dataset.
      On SEL~\cite{lee2017semantic} dataset, our method (without ER) significantly outperforms other competitors in terms of
      average F-measure.
      `CD,' `EMD,' and `EA' are different evaluation metrics described in \cref{sec:metric}.
    }\vspace{-6pt}
    \begin{tabular}{l|c|ccc|ccc|ccc}
    \toprule
    \multirow{2}{*}{Dataset} & \multirow{2}{*}{Method} & \multicolumn{3}{c|}{CD} & \multicolumn{3}{c|}{EMD} & \multicolumn{3}{c}{EA} \\
    & & Avg. P & Avg. R & Avg. F & Avg. P & Avg. R & Avg. F & Avg. P & Avg. R & Avg. F \\
    \hline
    \multirow{6}{*}{SEL~\cite{lee2017semantic}} &
    HED~\cite{xie2015holistically} + HT~\cite{duda1971use} & 0.491 & 0.578 & 0.531 & 0.461 & 0.543 & 0.498 & 0.356 & 0.420 & 0.385 \\
    & SLNet-iter1~\cite{lee2017semantic} & 0.740 & \textbf{0.905} & 0.812  & 0.723 & \textbf{0.888} & 0.797 & 0.654 & \textbf{0.803} & 0.721 \\
    & SLNet-iter5~\cite{lee2017semantic} & 0.826 & 0.841 & 0.834 & 0.810 & 0.824 & 0.817 & 0.735 & 0.747 & 0.741  \\
    & SLNet-iter10~\cite{lee2017semantic} & 0.858 & 0.821 &  0.839 & 0.840 & 0.804 & 0.822 &0.762 & 0.729 & 0.745 \\
    & Ours (VGG16) & 0.841 & 0.835 & 0.838 & 0.830 & 0.824 & 0.827 & 0.756 & 0.774 & 0.765\\
    & Ours (ResNet50) & \textbf{0.886} & 0.815 & \textbf{0.849} & \textbf{0.878} & 0.807 & \textbf{0.841} & \textbf{0.819} & 0.755 & \textbf{0.786} \\ 
    \hline
    \multirow{3}{*}{NKL} &
    HED~\cite{xie2015holistically} + HT~\cite{duda1971use} & 0.301 & 0.878 & 0.448 & - & - & - & 0.213 & 0.622 & 0.318 \\
    & Ours (VGG16) & 0.750 & 0.864 & 0.803 & 0.726 & 0.837 & 0.778 & 0.659 & 0.759 & 0.706 \\
    & Ours (ResNet50) & 0.766 & 0.864 & 0.812 & 0.743 & 0.839 & 0.789 & 0.679 & 0.766 & 0.719 \\ 
    \bottomrule
    \end{tabular}
    \label{tab:quantitative}
\end{table*}
}
\CheckRmv{
\begin{figure*}[!htb]
  \begin{center}
  \begin{overpic}[width=0.9\linewidth]{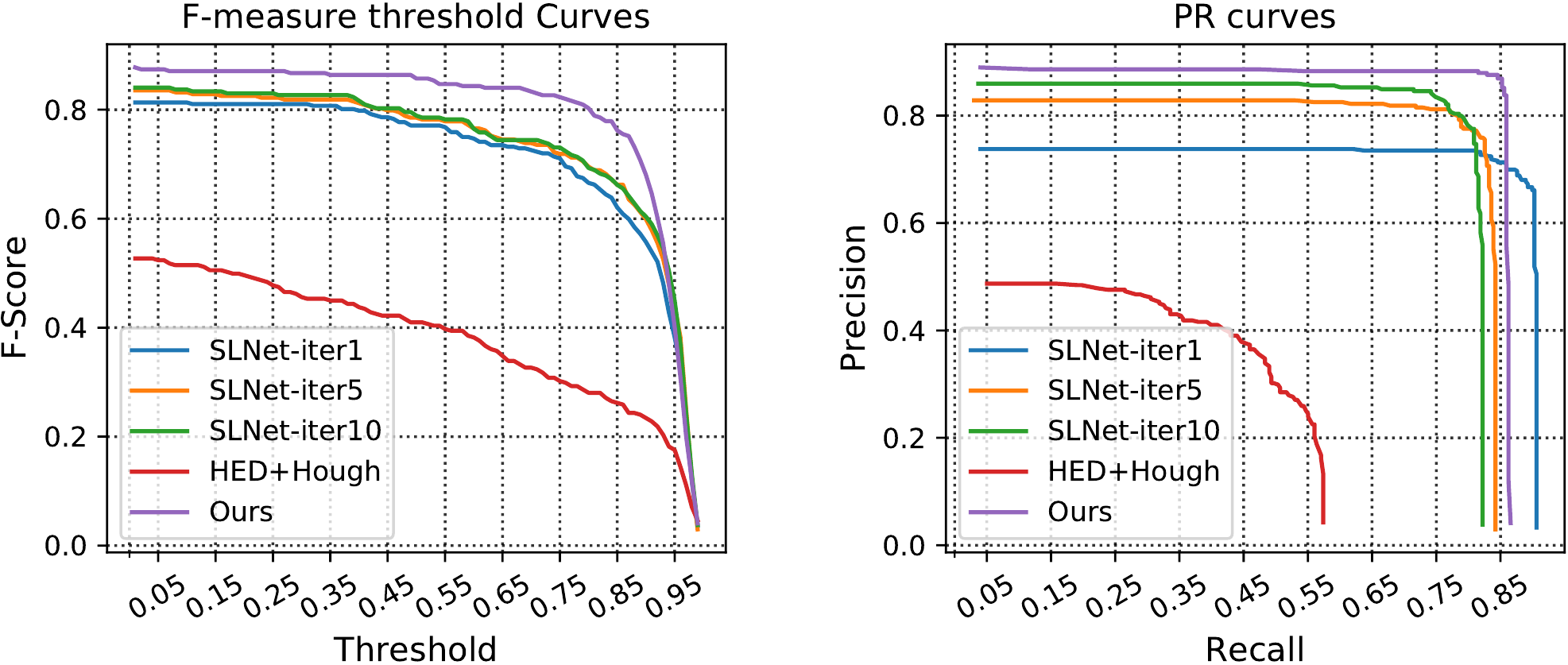}
  \end{overpic}
  \end{center}\vspace{-10pt}
  \caption{
    Left: F-measure under various thresholds.
    Right: The precision-recall curve.
    Out method outperforms SLNet~\cite{lee2017semantic} and classical
    Hough transform~\cite{duda1971use} with a
    considerable margin.
    Moreover, even with 10 rounds of location refinement,
    SLNet still presents inferior performance.
  }\vspace{20pt}
  \label{fig:curves}
\end{figure*}
}

\CheckRmv{
\begin{figure*}[htb!]
  \begin{center}
    \begin{overpic}[width=1.0 \linewidth]{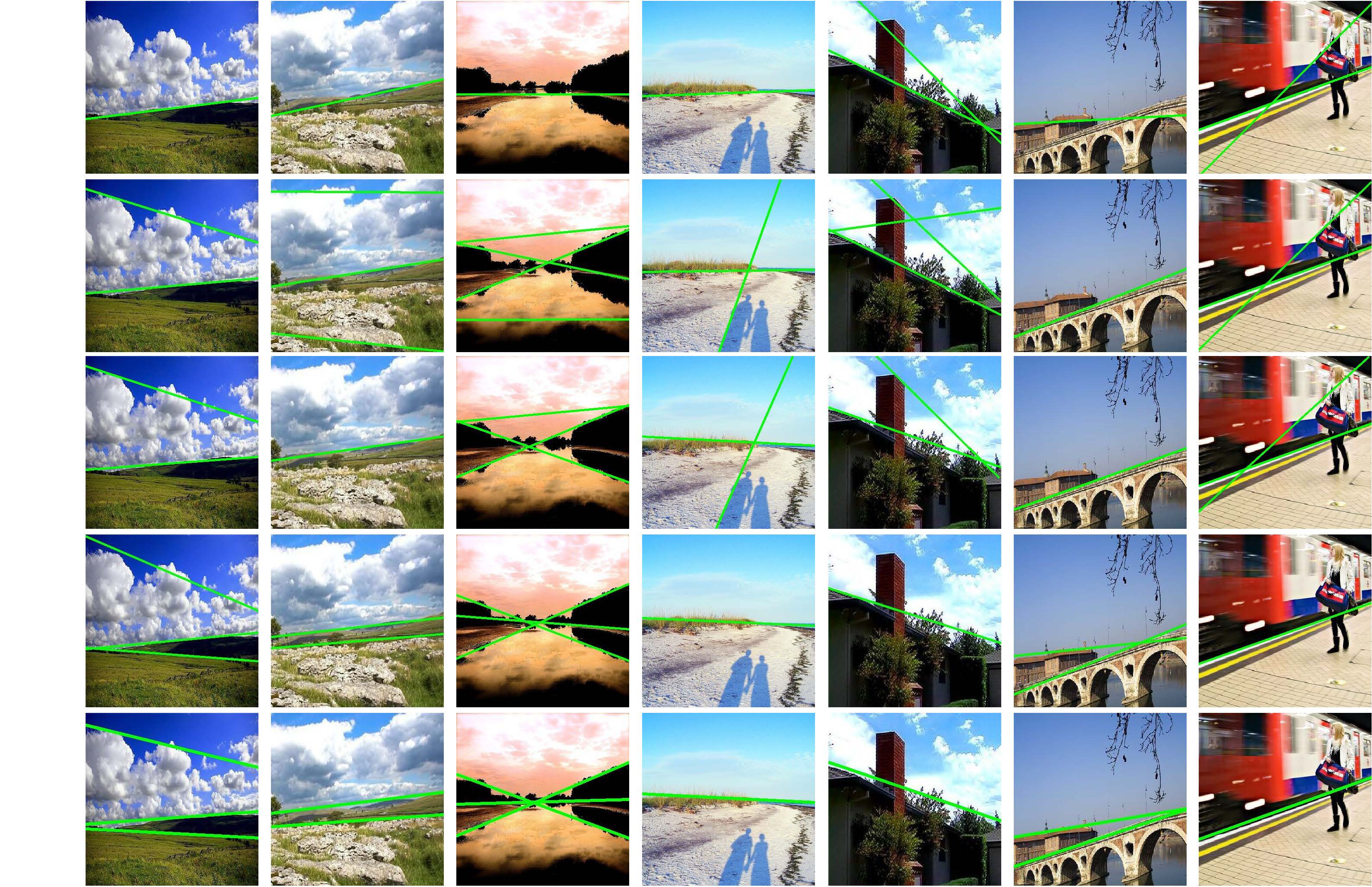}
      \put(0.5, 5){GT}
      \put(0, 19){Ours}

      \put(-1, 34.5){SLNet}
      \put(-1.3, 32){(iter10)}

      \put(-1, 47){SLNet}
      \put(-0.5, 44.5){(iter1)}


      \put(-1, 59){HED}
      \put(-1, 56.5){+ HT}
  \end{overpic}
  \end{center}
  \vspace{-12pt}
  \caption{
    Example detection results of different methods on the SEL dataset.
    Compared to SLNet~\cite{lee2017semantic}
    and classical Hough transform~\cite{duda1971use},
    our results are more consistent with the ground-truth.
  }
  \label{fig:detections-SEL}
\end{figure*}
}

\subsubsection{Results on the \revise{NKL} dataset}
We report the performance of our newly constructed \revise{NKL} dataset.
Since SLNet~\cite{lee2017semantic} did not release the training code,
we only compare our method with HED+HT.
\rerevise{As shown in ~\cref{tab:quantitative}, our proposed method outperforms
the baseline method (HED edge detector + Hough transform) with a clear margin.}


\CheckRmv{
\begin{figure*}[tb!]
  \begin{center}
  \begin{overpic}[width=1.0 \linewidth]{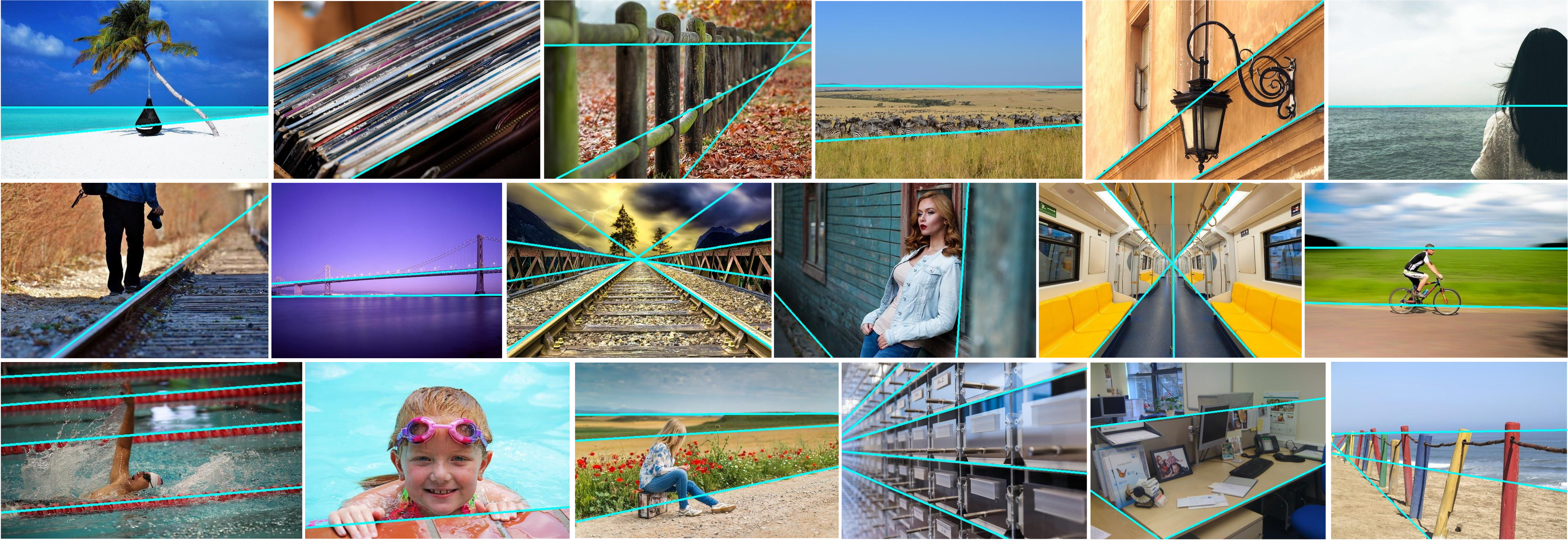}
  \end{overpic}
  \end{center}
  \vspace{-12pt}
  \caption{
    Detection results of our method on the \revise{NKL} dataset.
    Our method produces results that are visually compatible with human perception.
  }
  \label{fig:detections-NKL}
\end{figure*}
}

\subsubsection{Runtime efficiency.}
In this section, we benchmark the runtime of different methods including SLNet~\cite{lee2017semantic}
with various iteration steps, classical Hough transform and our proposed method.

Both SLNet~\cite{lee2017semantic} and HT require HED~\cite{xie2015holistically}
edge detector as a preprocessing step.
The non-maximal suppression (NMS) in SLNet requires edge maps as guidance,
and the classical Hough transform takes an edge map as input.
Moreover, SLNet uses a refining network to enhance the results iteratively.
Therefore, the inference speed is related to the iteration  steps.
In contrast, our method produces output results with a single forward pass,
and the NMS is as simple as computing the centroids of each connected area
in the parametric space.

Results in~\cref{tab:speed} illustrate that our method is significantly faster than
all other competitors with a very considerable margin.
Even with only 1 iteration step, SLNet is still slower than our method.

\CheckRmv{
\begin{table}[!htb]
  \renewcommand{\arraystretch}{1.4}
  \renewcommand\tabcolsep{4.4pt}
  \centering
  \caption{
      Quantitative speed comparisons.
      Our method \revise{(without ER)} is much faster than the other two competitors in network forward.
      Furthermore, our method doesn't require any extra-process \eg edge detection.
      As a result, our method can run at 49 FPS, which is remarkably
      higher than the other two methods.
  }\vspace{-6pt}
  \label{tab:speed}
  \begin{tabular}{l|c|c|c|c}
  \toprule
  Method & Network forward & NMS & Edge & Total \\
  \hline
  SLNet-iter1~\cite{lee2017semantic} & 0.354\ s & 0.079\ s & 0.014\ s & 0.447\ s \\
  SLNet-iter3~\cite{lee2017semantic} & 0.437\ s & 0.071\ s & 0.014\ s & 0.522\ s \\
  SLNet-iter10~\cite{lee2017semantic} & 0.827\ s & 0.068\ s & 0.014\ s & 0.909\ s \\
  HED~\cite{xie2015holistically} + HT~\cite{duda1971use} & 0.014\ s & 0.117\ s & 0.024\ s &  0.155\ s \\
  \hline
  Ours (VGG16) & 0.03\ s & 0.003\ s & 0 & 0.033\ s \\
  \hline
  Ours (ResNet50) & 0.017\ s & 0.003\ s & 0 & \textbf{0.020\ s} \\
  \bottomrule
  \end{tabular}
\end{table}
}

\subsection{Qualitative Comparisons } \label{sec:visual}
Here we give several example results of our proposed method along with SNLet
and HED+HT.
As shown in~\cref{fig:detections-SEL}, compared with other methods, our results are more compatible with
the ground-truth as well as human cognition.
In addition to the results in~\cref{fig:detections-SEL}, we provide all the detection results of our method
and SLNet \revise{in the supplementary materials}.

\subsection{Ablation Study} \label{sec:ablation}
In this section, we ablate each of the components in our method.

\subsubsection{Components in DHT} \label{sec:ablation-dht}
We first ablate components of ``deep Hough transform''.
Specifically, they are:
(a) the Deep Hough transform (DHT) module detailed in~\cref{sec:dht-dht};
(b) the multi-scale (MS) DHT architecture described in~\cref{sec:ms-dht-fpn};
(c) the context-aware (CTX) line detector proposed in~\cref{sec:ctx-line-detector}.
Experimental results are shown in~\cref{tab:ablation}.

We first construct a baseline model with plain ResNet50 and DHT module.
%
%
Then we verify the effectiveness of the multi-scale (MS) strategy and context-aware
line detector (CTX), individually.
We separately append MS and CTX to the baseline model and then evaluate their performance, respectively.
Results in~\cref{tab:ablation} indicate that both MS and CTX can improve
the performance of the baseline model.

At last, we combine all the components to form our final full method,
which achieves the best performance among all other combinations.
Experimental results in this section clearly demonstrate that each component of
our proposed method contributes to the success of our method.

\CheckRmv{
\begin{table}[!htb]
    \renewcommand{\arraystretch}{1.3}
    \renewcommand\tabcolsep{1.0pt}
    \newcolumntype{C}{>{\centering\arraybackslash}p{0.11\textwidth}}
    \centering
    \caption{
        Ablation study for each component. MS indicates DHTs with multi-scale features as described in~\cref{sec:ms-dht-fpn},
        and CTX means context-aware aggregation as described in~\cref{sec:ctx-line-detector}.
    }\vspace{-6pt}
    \begin{tabular}{C|C|C|C}
    \toprule
    DHT & MS & CTX & F-measure \\
    \hline
    \checkmark  &  &  & 0.664 \\
    \checkmark & \checkmark &  & 0.758 \\
    \checkmark & & \checkmark & 0.771 \\
    \checkmark & \checkmark & \checkmark & 0.786 \\
    \bottomrule
    \end{tabular}
    
    \label{tab:ablation}
\end{table}
}

\subsubsection{Edge-guided Refinement} \label{sec:ablation-refinement}
Here we ablate the ``Edge-guided Refinement'' module \revise{(abbreviated as ER)}.
First, we test the performance of DHT+ER using different $\delta_r$.
The $\delta_r$ parameter controls the size of the searching space
in ER ($\mathcal{L}$ in ~\cref{eq:refine-search}).
This experiment is conducted on the SEL dataset using the ResNet50 backbone.
\CheckRmv{
\begin{table}[!htb]
  \renewcommand{\arraystretch}{1.3}
  \newcolumntype{C}{>{\centering\arraybackslash}p{0.08\textwidth}}
  \centering
  \caption{
    Performance DHT+ER with different $\delta_r$.
    Models are trained/tested on the SEL dataset using the Resnet50 backbone.
    $\delta_r=0$ represents with vanilla DHT method without ER.
  }\vspace{-6pt}
  \newcommand{\CC}{\cellcolor{gray!20}}
  \begin{tabular}{C|C|C|C}
  \toprule
  $\delta_r$ & Precision & Recall & F-measure\\
  \hline
  \CC 0 & \CC 0.8190 & \CC 0.7530 & \CC 0.7861 \\
  1 & 0.8199 & 0.7561 & 0.7866 \\
  3 & 0.8208 & 0.7569 & 0.7874 \\
  5 & 0.8214 & 0.7574 & 0.7880 \\
  7 & 0.8213 & 0.7573 & 0.7878 \\
  9 & 0.8212 & 0.7571 & 0.7877 \\
  \bottomrule
  \end{tabular}
  \label{tab:ablation-refinement-1}
\end{table}
}
Results in ~\cref{tab:ablation-refinement-1} tells that the performance first increases and
then gets saturated with the growth of $\delta_r$.
Since the peak performance occurs when $\delta_r = 5$,
we set $\delta_r=5$ for better performance.
After setting $\delta_r$ to 5, we compare the performance of our method with and without
ER, using different backbones \revise{and} datasets.

\CheckRmv{
\begin{table}[!htb]
  \renewcommand{\arraystretch}{1.3}
  \centering
  \caption{
    Performance with and without ER ($\delta_r=5$) using different backbones \revise{and} datasets.
  }\vspace{-6pt}
  \newcommand{\CC}{\cellcolor{gray!20}}
  \begin{tabular}{l|c|c|c|c|c|c}
  \toprule
  Dataset & Arch & Edge & P & R & F & F@0.95\\
  \hline
  \multirow{4}{*}{SEL~\cite{lee2017semantic}} & VGG16 &  & 0.756 & 0.774 & 0.765 & 0.380\\
   & \CC VGG16  & \CC \checkmark & \CC 0.758 & \CC 0.777 & \CC 0.770 & \CC 0.439 \\
   & Resnet50 & & 0.819 & 0.753 & 0.786 & 0.420\\
   & \CC Resnet50 & \CC \checkmark & \CC 0.821 & \CC 0.757 & \CC 0.788 & \CC 0.461\\
   \hline
   \multirow{4}{*}{\revise{NKL}} & VGG16 &   & \revise{0.659} & \revise{0.759} & \revise{0.706} & \revise{0.434}\\
    & \CC VGG16 & \CC\checkmark & \CC \revise{0.664} & \CC \revise{0.765} & \CC \revise{0.711} & \CC \revise{0.472}\\
    & Resnet50 &  & \revise{0.679} & \revise{0.766} & \revise{0.719} & \revise{0.459}\\
    & \CC Resnet50 & \CC \checkmark & \CC \revise{0.684} & \CC \revise{0.771} & \CC \revise{0.725} & \CC \revise{0.486}\\
  \bottomrule
  \end{tabular}
  \label{tab:ablation-refinement-2}
\end{table}
}

Results in ~\cref{tab:ablation-refinement-2} clearly demonstrate that
edge-guided refinement can effectively improve detection results regardless
of backbone architectures and datasets.

\section{Conclusions}\label{sec:conclusion}
In this paper, we proposed a simple yet effective method for semantic line detection in
natural scenes.
By incorporating the strong learning ability of CNNs into classical Hough transform,
our method is able to capture complex textures and rich contextual semantics of lines.
To better assess the similarity between a pair of lines,
we designed a new evaluation metric considering both Euclidean distance and angular distance
between lines.
Besides, a new dataset for semantic line detection
was constructed to fulfill the gap between the scale of existing datasets
and the complexity of modern CNN models.
Both quantitative and qualitative results revealed that
our method significantly outperforms previous arts in terms of both detection quality and speed.

\section*{Acknowledgment}
This research was supported by the National Key Research and 
Development Program of China (2018AAA0100400), 
NSFC (61922046,61620106008,62002176), 
S\&T innovation project from Chinese Ministry of Education,
and Tianjin Natural Science Foundation (17JCJQJC43700).

\bibliographystyle{IEEEtran}
\bibliography{line}

\ifCLASSOPTIONcaptionsoff
  \newpage
\fi
\newcommand{\AddPhoto}[1]{\includegraphics%
[width=1in,height=1.25in,clip,keepaspectratio]{figures/photos/#1}}

\vfill

\end{document}


\title{Supplementary Material for `Deep Hough Transform for Semantic Line Detection'}

\author{\IEEEauthorblockN{Kai Zhao\thanks{\IEEEauthorrefmark{1} The first two students contribute equally to this paper.}\IEEEauthorrefmark{1}},
        Qi Han\IEEEauthorrefmark{1},
         Chang-Bin Zhang,
         Jun Xu,
         \IEEEauthorblockN{Ming-Ming Cheng\thanks{\IEEEauthorrefmark{2} M.M. Cheng is the corresponding author (cmm@nankai.edu.cn).}\IEEEauthorrefmark{2}},~\IEEEmembership{Senior Member,~IEEE}

 \IEEEcompsocitemizethanks{
   \IEEEcompsocthanksitem The authors are with the TKLNDST, 
     College of Computer Science, Nankai University, Tianjin, China, 300350.
   \IEEEcompsocthanksitem A preliminary version of this work has been presented in \cite{eccv2020line}.
 }
 \thanks{Manuscript received August 11, 2020. 
 }}

\twocolumn[{%
\renewcommand\twocolumn[1][]{#1}%
\maketitle
\begin{center}
    \centering
    \begin{overpic}[width=1 \linewidth]{arch_details.pdf}
      \put(0.7, 13.2){Stage1}
      \put(0.7, 23){Stage2}
      \put(0.7, 32.4){Stage3}
      \put(0.7, 41.6){Stage4}
      \put(39.4, 13){$\mathcal{H}$}
      \put(39.4, 22.2){$\mathcal{H}$}
      \put(39.4, 32.2){$\mathcal{H}$}
      \put(39.4, 41.6){$\mathcal{H}$}
      \put(78.8, 31.8){$\mathcal{R}$}
      \put(76, 14){Add}
      \put(88, 14){$\uparrow$ Concat}
      \put(3.5,2.5){\scriptsize Conv$3\times3$, BN, ReLU}
      \put(31,2.5){\scriptsize Conv$1\times1$, BN, ReLU}
      \put(59,2.5){\scriptsize Conv$1\times1$}
      \put(59,2.5){\scriptsize Conv$1\times1$}
      \put(73.2,2.5){$\mathcal{H}$}
      \put(77,2.5){\scriptsize DHT}
      \put(89,2.5){$\mathcal{R}$}
      \put(93,2.5){\scriptsize RHT}
    \end{overpic}
    \label{fig:net-arch}
\end{center}%
}]

%
In the supplementary material, we provide
(1) the detailed network architectures,
and (2) additional experimental results.

\section{Network Architecture}\label{sec:net-arch}
Here we detail the proposed network architectures.
The proposed Deep Hough Transform (DHT) module and Reverse Hough Transform (RHT) are embeded into
a FPN backbone.
Taking ResNet50 based FPN as an example,  details are given in the above figure.

\begin{figure*}[t]
  \centering
  \includegraphics[width=.95\linewidth]{result_00.jpg}
\end{figure*}

\begin{figure*}[t]
  \centering
  \includegraphics[width=.95\linewidth]{result_11.jpg}
\end{figure*}

\begin{figure*}[!t]
  \centering
  \includegraphics[width=1\linewidth]{result_22.jpg}
\end{figure*}

\section{All Detection Results on the SEL Dataset}

We provide ALL the detection results on the SEL dataset \cite{lee2017semantic}.
Bottom rows are result of SLNet and upper rows are results
produced by our method.

\bibliographystyle{IEEEtran}
\bibliography{line}